\pdfoutput=1
\documentclass{article}

\usepackage{microtype}
\usepackage{graphicx}
\usepackage{subfigure}
\usepackage{booktabs} 

\usepackage{hyperref}

\usepackage[accepted]{icml2024}

\usepackage{amsmath}
\usepackage{cancel} 
\usepackage{amssymb}
\usepackage{mathtools}
\usepackage{amsthm}
\usepackage{bbm}
\usepackage{xcolor}

\usepackage[capitalize,noabbrev]{cleveref}

\theoremstyle{plain}
\newtheorem{theorem}{Theorem}[section]
\newtheorem{proposition}[theorem]{Proposition}

\theoremstyle{definition}
\newtheorem{definition}[theorem]{Definition}

\theoremstyle{remark}

\usepackage[textsize=tiny]{todonotes}

\icmltitlerunning{Model Alignment as Prospect Theoretic Optimization}

\begin{document}

\twocolumn[
\icmltitle{KTO: Model Alignment as Prospect Theoretic Optimization}



\icmlsetsymbol{equal}{*}

\begin{icmlauthorlist}
\icmlauthor{Kawin Ethayarajh}{stanford}
\icmlauthor{Winnie Xu}{contextual}
\icmlauthor{Niklas Muennighoff}{contextual}
\icmlauthor{Dan Jurafsky}{stanford}
\icmlauthor{Douwe Kiela}{stanford,contextual}
\end{icmlauthorlist}

\icmlaffiliation{stanford}{Stanford University (first author was an intern at Contextual AI)}
\icmlaffiliation{contextual}{Contextual AI}

\icmlcorrespondingauthor{Kawin Ethayarajh}{kawin@stanford.edu}

\icmlkeywords{Machine Learning, ICML}

\vskip 0.3in
]



\printAffiliationsAndNotice{}  

\begin{abstract}
Kahneman \& Tversky's \emph{prospect theory} tells us that humans perceive random variables in a biased but well-defined manner \citeyearpar{tversky1992advances}; for example, humans are famously loss-averse.
We show that objectives for aligning LLMs with human feedback implicitly incorporate many of these biases---the success of these objectives (e.g., DPO) over cross-entropy minimization can partly be ascribed to them belonging to a family of loss functions that we call \emph{human-aware losses} (HALOs).
However, the utility functions these methods attribute to humans still differ from those in the prospect theory literature.
Using a Kahneman-Tversky model of human utility, we propose a HALO that directly maximizes the utility of generations instead of maximizing the log-likelihood of preferences, as current methods do. 
We call this approach KTO, and it matches or exceeds the performance of preference-based methods at scales from 1B to 30B, despite only learning from a binary signal of whether an output is desirable.
More broadly, our work suggests that there is no one HALO that is universally superior; the best loss depends on the inductive biases most appropriate for a given setting, an oft-overlooked consideration.
\end{abstract}

\section{Introduction}

Aligning generative models with human feedback has been successfully used to make generations more helpful, factual, and ethical, among other desiderata \citep{ouyang2022training,tian2023fine}.
For LLMs, alignment methods such as RLHF and DPO have consistently proven to be more beneficial than doing supervised finetuning (SFT) alone. 
However, human feedback is often discussed only in the context of preferences (e.g., output $y_w \succ y_l$ for input $x$), even though it can take many forms (e.g., approval/disapproval of $y$ given $x$).
This is because preferences, despite being a kind of data that is relatively scarce and expensive to collect in practice \citep{casper2023open}, are required by the alignment methods shown to work best---RLHF~\citep{christiano2017deep} and DPO~\citep{rafailov2023direct}.

To understand why these methods work so well, and whether feedback needs to be in preference form, we frame alignment through the lens of \textit{prospect theory}~\cite{kahneman1979prospect,tversky1992advances}.
Prospect theory explains why humans make decisions about uncertain events that do not maximize their expected value.
It formalizes how humans perceive random variables in a biased but well-defined manner; for example, relative to some reference point, humans are more sensitive to losses than gains, a property called \textit{loss aversion}.
We show that popular alignment methods such as DPO and PPO-Clip \citep{schulman2017proximal} implicitly model some of these biases, helping explain their success independently of the data used (\S\ref{ssec:halos}).
We then propose a more general class of such loss functions called \textit{human-aware losses} (HALOs).\footnote{We use the term \emph{human-aware} to draw an analogy with how \emph{hardware-aware} methods benefit from being designed around hardware limitations \citep{dao2022flashattention}, not to claim that a simple loss function is fully aware of human behavior.}

\begin{figure}
    \centering
    \includegraphics[width=0.9\columnwidth]{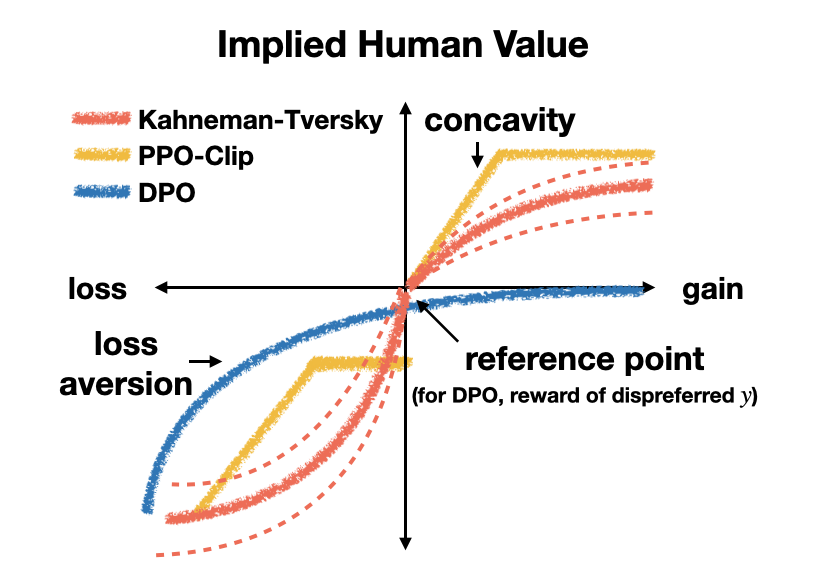}
    \vspace{-5pt}
    \caption{The utility that a human gets from the outcome of a random variable, as implied by different human-aware losses (HALOs).
    Notice that the implied value functions share properties such as loss aversion with the canonical human value function in prospect theory \citep{tversky1992advances}.
     }
    \label{fig:utility}
\end{figure}

Although it is impossible to say that HALOs are categorically better than non-HALOs, we find that among existing methods, those that meet the definition of a HALO work better than those that do not (\S\ref{ssec:do_halos_matter}).
We find that DPO performance can even be matched at most scales by running an offline PPO variant on dummy +1/-1 rewards, suggesting that preference data might not be needed if the inductive bias in the loss function is good enough.
However, despite the surprising success of this simple baseline, it significantly lags behind DPO at the 30B LLM scale and suffers from hyperparameter sensitivity, making it difficult to use.

Taking a more principled approach, we derive a HALO using the model of human utility that Kahneman \& Tversky proposed to describe how humans make decisions about uncertain monetary outcomes \citep{tversky1992advances}.
This approach, which we call Kahneman-Tversky Optimization (KTO), directly maximizes the utility of generations instead of maximizing the log-likelihood of preferences, as most current methods do (\S\ref{ssec:kto}).
KTO only requires a binary signal of whether an output is desirable or undesirable for an input.
This data is more abundant, cheaper, and faster to collect in the real world,  making it easier to scale alignment in production and rapidly iterate on models.
We find that:
\begin{itemize}
    \item KTO matches or exceeds DPO performance at  scales from 1B to 30B parameters (\S\ref{ssec:experiments}).\footnote{Our code is available on \href{https://github.com/ContextualAI/HALOs}{Github}; models are on \href{https://huggingface.co/collections/ContextualAI/archangel-65bd45029fa020161b052430}{Hugging Face}.}
    That is, breaking up $n$ preferences meant for DPO into $2n$ examples for KTO can yield better LLM generations, as determined by closed-ended tasks such as mathematical reasoning and open-ended judgments from humans and \texttt{GPT-4}. 
    
    \item KTO can handle extreme data imbalances, matching DPO performance while using up to 90\% fewer desirable examples (i.e., examples of good generations).
    Its success cannot be ascribed to the alignment data being sourced from a preference dataset.

    \item When the pretrained model is sufficiently good, one can skip supervised finetuning (SFT) and go straight to KTO without a loss in generation quality, whereas SFT is always needed for best results with DPO.
\end{itemize}
The intent behind KTO was that even if the model learns from a weaker signal, we could compensate with the higher volume of data that could be accessed in practice; the fact that KTO can match and even outperform DPO on the same data is thus surprising.
We conclude by discussing some theoretical explanations for this phenomenon (\S\ref{ssec:theory}).
Despite the success of KTO in our experiments, our work ultimately suggests that there is no one HALO that is universally superior; the best HALO depends on the inductive biases appropriate for a given setting, and this choice should be made deliberately instead of defaulting to any one loss.

\section{Background}
\label{sec:background}
For a detailed survey, please see Appendix \ref{sec:appendix_related_work}.
In brief, LLMs are traditionally trained in three stages \citep{ouyang2022training}:

\paragraph{Pretraining} Given a large corpus, train the model to maximize the log-likelihood of the next token conditioned on the preceding text.
Let $\pi_0$ denote the pretrained model.

\paragraph{Supervised Finetuning (SFT)} Finetune the model to predict the next token on data that is more relevant to the downstream task.
Often, such data will comprise instructions and an appropriate response (i.e., instruction finetuning).
Let $\pi_\text{ref}$ denote the finetuned model.

\paragraph{RLHF} Given a dataset $\mathcal{D}$ of preferences $(x, y_w, y_l)$---where $x$ is an input, $y_w, y_l$ are the preferred and dispreferred outputs (i.e., $y_w \succ y_l$ for $x$), and $r^*$ is the “true” reward function underlying the preferences---it is first assumed that the probability that $y_w$ is preferred to $y_l$ can be captured with a specific function class, typically a Bradley-Terry model \citep{bradley1952rank}. Where $\sigma$ is the logistic function:
\begin{equation}
p^*(y_w \succ y_l|x) = \sigma(r^*(x,y_w) - r^*(x,y_l))
\label{eq:bradley}
\end{equation}
Since getting the true reward from a human would be intractably expensive, a reward model $r_\phi$ learns to serve as a proxy, done by minimizing the negative log-likelihood of the human preference data:
$$\mathcal{L}_R(r_\phi) = \mathbb{E}_{x,y_w,y_l \sim D}[- \log \sigma(r_\phi(x, y_w) - r_\phi(x, y_l))]$$

But solely maximizing the reward might come at the expense of desiderata such as generating grammatical text. 
To avoid this, a KL divergence penalty is introduced to restrict how far the language model can drift from $\pi_\text{ref}$. 
Where $\pi_\theta$ is the model we are optimizing, the optimal model $\pi^*$ is that which maximizes
\begin{equation}
\begin{split}
    \mathbb{E}_{x \in D, y \in \pi_\theta} [r_\phi(x,y)] \
    & - \beta D_{\text{KL}}(\pi_\theta(y|x) \| \pi_{\text{ref}}(y|x)) 
    \label{eq:rlhf_obj}
\end{split}
\end{equation}
where $\beta > 0$ is a hyperparameter. 
This objective is typically optimized using a policy-gradient algorithm such as PPO \citep{schulman2017proximal}.

However, RLHF is often slow (largely because of having to sample generations) and quite unstable in practice (especially in a distributed setting).
For this reason, recent work has focused on designing closed-form losses that maximize the margin between the preferred and dispreferred generations.
In particular, Direct Preference Optimization (DPO) \citep{rafailov2023direct} has emerged as a popular alternative as it allows the same optimal policy as in RLHF to be recovered under certain conditions:
\begin{equation}
\begin{split}
    & \mathcal{L}_\text{DPO}(\pi_\theta, \pi_\text{ref}) = \mathbb{E}_{x,y_w,y_l \sim D} \\ & \left[  -\log \sigma \left( \beta \log \frac{\pi_\theta(y_w|x)}{\pi_\text{ref}(y_w|x)} - \beta \log \frac{\pi_\theta(y_l|x)}{\pi_\text{ref}(y_l|x)} \right) \right]
\end{split}
\label{eq:dpo}
\end{equation}

\section{A Prospect Theoretic View of Alignment}

To understand why alignment methods work so well, we now frame them through the lens of \textit{prospect theory} \citep{tversky1992advances}.
Prospect theory explains why, when faced with an uncertain event, humans make decisions that do not maximize their expected value.
For example, because humans are loss-averse, given a gamble that returns \$100 with 80\% probability and \$60 with 20\% probability, a person might accept \$60 to avoid the gamble, despite their \textit{certainty equivalent} of \$60 being less than the expected value of \$80.

\subsection{Prospect Theory}

In prospect theory, human utility depends on a \textit{value function} and a \textit{weighting function}:\footnote{\textit{Cumulative prospect theory} is the full name of the expanded theory we discuss here \citep{tversky1992advances}.}

\begin{definition}
A \textit{value function} $v: \mathcal{Z} \to \mathbb{R}$ maps an outcome $z$, relative to some reference point $z_0$, to its perceived (or subjective) value.
For example, these functions capture the fact that humans tend to be more sensitive to relative losses than relative gains of the same magnitude.
\end{definition}

\begin{definition}
A \textit{weighting function} $\omega$ is the derivative of a \textit{capacity function} that maps cumulative probabilities to perceived cumulative probabilities.
These functions capture, for example, the fact that humans tend to overestimate the chance of rare events.
Let $\omega_z$ denote the weight placed on outcome $z$.
\end{definition}

\begin{definition}
The \textit{utility of a random variable} $Z$ is a function of its outcomes: $u(Z) \triangleq \sum_{z \in Z} \omega_z v(z; z_0)$.
\end{definition}

However, because humans do not see the full probability distribution of an LLM, weighting functions are not salient to this discussion; we will focus only on value functions.
Using experiments that presented real humans with monetary gambles and asked for their certainty equivalent, \citet{tversky1992advances} proposed the following functional form for human value:
\begin{equation}
    v(z; \lambda, \alpha,  z_0) = \begin{cases}
(z - z_0)^\alpha & \text{if } z \geq z_0\\
-\lambda(z_0 - z)^\alpha & \text{if } z < z_0 \\
    \end{cases}
    \label{eq:human_value}
\end{equation}
where the median value of hyperparameter $\alpha = 0.88$ and $\lambda = 2.25$ across individuals.
$\alpha$ controls the curvature of the function, which reflects risk aversion; $\lambda$ controls its steepness, which reflects loss aversion.
While the shape of the median Kahneman-Tversky value function is illustrated in Figure \ref{fig:utility}, it should be noted that it varies across individuals \citep{tversky1992advances}.
There are also other functional forms for the value function that have been proposed in later work \citep{gurevich2009decision}. 
The salient qualities of a value function are: the existence of a reference point that is used to get the \emph{relative} gain or loss; concavity in relative gains (i.e., diminishing sensitivity away from $z_0$); and loss aversion (i.e., greater sensitivity to losses).

\subsection{HALOs}
\label{ssec:halos}

\begin{definition}[\textbf{HALOs}]
    Let $\theta$ denote the trainable parameters of the model $\pi_\theta: \mathcal{X} \to \mathcal{P}(\mathcal{Y})$ being aligned, $\pi_\text{ref}$ the reference model, $l: \mathcal{Y} \to \mathbbm{R}^+$ a normalizing factor, and $r_\theta(x,y) = {l(y)} \log [\pi_\theta(y|x) / \pi_\text{ref}(y|x)]$ the implied reward.
    Where $Q(Y'|x)$ is a reference point distribution over $\mathcal{Y}$  and $v_{x,y}: \mathbbm{R} \to \mathbbm{R}$ is non-decreasing everywhere, the \textit{human value} of $(x,y)$ is 
    \begin{equation}
        v_{x,y}(r_\theta(x,y) - \mathbb{E}_{Q}[r_\theta(x,Y')])
    \end{equation}
    A function $f$ is a \textit{human-aware loss} for $v$ if $\exists\ a_{x,y} \in \mathbb{R} \backslash \{0\}$ such that:
    \begin{equation}
    \begin{split}
        f(& \pi_\theta, \pi_\text{ref}) = \\  & \mathbb{E}_{x,y\sim\mathcal{D}}[ a_{x,y} v_{x,y}(r_\theta(x,y) - \mathbb{E}_{Q}[r_\theta(x, Y')])] + C_\mathcal{D}
    \end{split}
    \end{equation}
    where $\mathcal{D}$ is the feedback data and $C_\mathcal{D} \in \mathbb{R}$ is a data-specific constant. 
    Holding $\mathcal D$ fixed, the normalizing factor $l$, coefficients $a_{x,y}$, value functions $v_{x,y}$, and constant $C_{\mathcal D}$ must be chosen independently of $\pi_\theta$ and $\pi_{\text{ref}}$; the reference-point distribution $Q$ may depend on either policy.
    \label{def:halo}
\end{definition}
In a classic prospect theory experiment, $r_\theta$ would be the dollar amount assigned to each outcome; here, $r_\theta$ is measured in nats, as the decrease in conditional surprisal when going from $\pi_\text{ref}$ to $\pi_\theta$, normalized according to $l$.
This follows naturally from the next-token prediction objective used to pretrain and finetune LLMs.
As $\pi_\theta$ is aligned, we would expect $r_\theta$ to grow increasingly positive for desirable outputs and increasingly negative for undesirable outputs.

Another perspective on the reward comes from the RLHF objective in (\ref{eq:rlhf_obj}).
The policy that maximizes this objective has a closed-form expression \citep{peng2019advantage,peters2007reinforcement}; where $Z(x)$ is the partition function:
$$\pi^*(y|x) = \frac{1}{Z(x)} \pi_{\text{ref}}(y|x) \exp\left(\frac{1}{\beta} r^*(x,y)\right)$$
Letting $l(\cdot) = \beta$, we get
\begin{equation}
    r_{\theta^*}(x, y) = r^*(x,y) - \beta \log Z(x)
    \label{eq:opt_reward}
\end{equation}
Under $\theta^*$, the HALO-defined reward is just the optimal reward shifted by an input-specific term, meaning that $r_{\theta^*}$ is in the same equivalence class as $r^*$ and would also induce the optimal policy $\pi^*$ (Lemma 1, \citet{rafailov2023direct}).

\begin{figure*}[t]
    \centering
    \includegraphics[width=\textwidth]{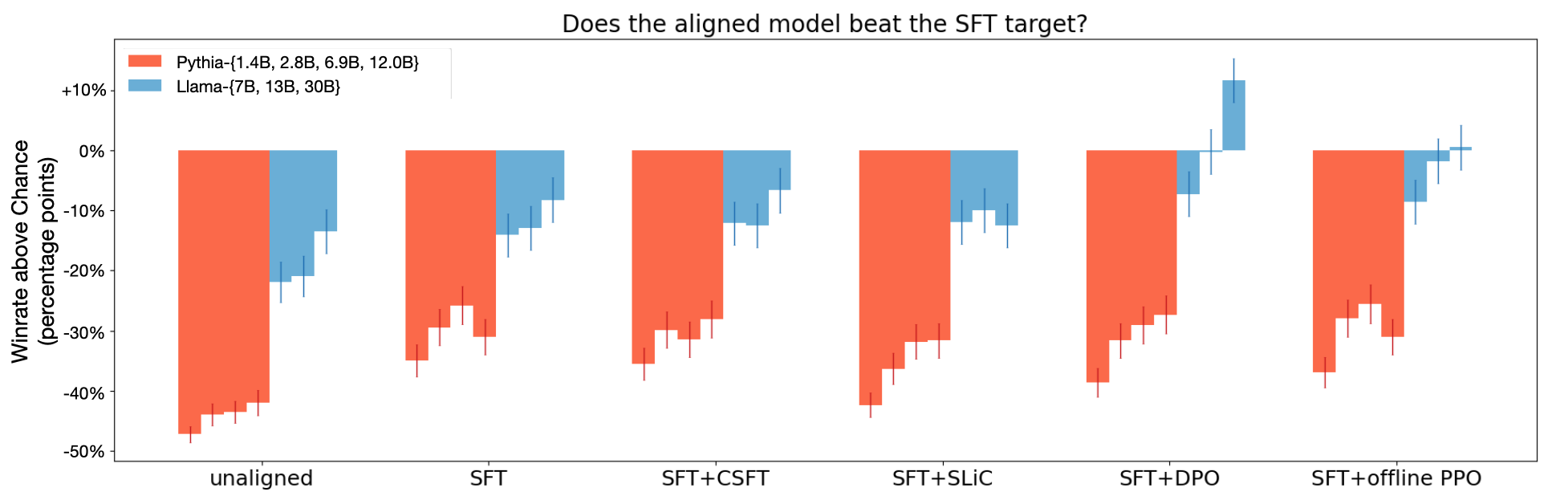}
    \vspace{-15pt}
    \caption{HALOs (DPO, offline PPO variant) outperform non-HALOs (SLiC, CSFT), as measured by the \texttt{GPT-4-0613}-judged winrate of the aligned model's generations against a hard-to-beat baseline: the outputs that would have been used as the targets for SFT.
    The $y$-axis here plots the winrate above chance (i.e., the winrate -- 50\%).
    The difference between methods is only significant $(p < 0.05)$ at 13B+ parameters, and only the HALO-aligned Llama-\{13B, 30B\} models are able to match the baseline and yield a winrate at or above chance.}
    \label{fig:halos_vs_nonhalos}
\end{figure*}

The reference point in a HALO is the expected reward from the human's perspective, where $Q(Y'|x)$ describes the examples that are used to construct a baseline.
We do not require that the value function be concave in gains or convex in losses however.
For one, a minority of individuals are risk-averse in both the gain and loss regime, resulting in concavity everywhere \citep{kahneman1979prospect}.
Second, the implied reward is already the output of a log function, so the value function may not be concave in this context.

\begin{theorem}
\label{theorem:halos}
    DPO and PPO-Clip are human-aware losses.
\end{theorem}
The proof is deferred to Appendix \ref{sec:appendix_proofs}. This can be seen more intuitively in Figure \ref{fig:utility}.

\subsection{Does being a HALO matter?}
\label{ssec:do_halos_matter}

It is difficult to answer whether being a HALO is useful, since both HALOs and non-HALOs are diverse function classes, but we attempt to do so by comparing popular methods that qualify as a HALO with those that do not:
\begin{enumerate}
    \item \textbf{CSFT}: Conditional SFT is a simple alignment method where a control token is prepended to the output during training; then, at inference, the control token corresponding to desirable generations is appended to the input to induce good generations \citep{korbak2023pretraining}.
    Note that for any choice of $\pi$, $f(\pi, \pi)$ should be a constant due to the invariance condition of HALOs.
    In contrast, the cross-entropy objective in CSFT generally varies with $\pi$---therefore CSFT is not a HALO.

    \item \textbf{SLiC}: Sequence Likelihood Calibration \citep{zhao2023slic}, which combines a max-margin loss for preferences with a language modeling loss, is not a HALO. 
    In particular, on setting
$\pi_\theta=\pi_{\text{ref}}=\pi$,
\[
\begin{aligned}
\mathcal L_{\text{SLiC}}(\pi,\pi)
&=
\mathbb E
\left[
\max\left(
0,
\delta-
\log
\frac{\pi(y_w\mid x)}
     {\pi(y_l\mid x)}
\right)
\right]
\\
&\quad+
\lambda_{\text{reg}}
\mathbb E_{x,y\sim\pi(\cdot\mid x)}
[-\log\pi(y\mid x)]
\end{aligned}
\]
which in general varies with $\pi$. This violates the invariance condition of HALOs.

    \item \textbf{DPO}: DPO is a HALO (Theorem \ref{theorem:halos}).

    \item \textbf{PPO (offline)}: The standard RLHF objective in (\ref{eq:rlhf_obj}) is typically optimized with PPO-Clip, which works by “clipping” how far $\pi_\theta$ can drift from the version $\pi_\text{old}$ at the previous step:
    \begin{equation*}
    \begin{split}
        \mathcal{L}_\text{PPO (offline)} =& -\mathbb{E}_{x,y,t \sim D}[\min(q_\theta A(x{:}y_{<t}, y_t), \\
        &\text{clip}(q_\theta, 1 - \epsilon, 1 + \epsilon) A(x{:}y_{<t}, y_t))]
    \end{split}
    \end{equation*}
    where $q_\theta = \frac{\pi_\theta (y_t|x{:}y_{<t})}{\pi_\text{old}(y_t|x{:}y_{<t})}$ and $A(x{:}y_{<t}, y_t)$ is the per-token advantage (i.e., the surplus benefit from producing a given token in a given state).
    
    PPO is an online algorithm---generations are sampled from the current model, judged by a reward model, and then used to update the current version. 
    However, for a fair comparison with offline methods, we consider a variant of PPO that is only fed offline data.
    Because RLHF is also quite unstable in a distributed setting, we never update $\pi_\text{old}$ and keep it as $\pi_\text{ref}$, instead clipping less conservatively than we traditionally would.
    \citet{baheti2023improving} found that these changes, along with treating the entire output sequence as a single action, greatly improves stability. 
    However, since RLHF traditionally uses token-level advantages, we omit the third change and only preserve the first two.
    The PPO-Clip loss is unchanged and remains a HALO (Theorem \ref{theorem:halos}).
    
    Calling this method PPO is somewhat imprecise, because it is offline and takes only one step, but to avoid introducing too many new terms, we will call this \textit{PPO (offline)}.
    Instead of using learned rewards, we simplify even further and use dummy +1/-1 rewards\footnote{Note that +1/-1 do not refer to the HALO-implied rewards, but rather the rewards used for the advantage $A(x{:}y_{<t}, y_t)$ calculation.} for $y_w$ and $y_l$ instead.
    Further details on the implementation of this method can be found in Appendix \ref{sec:appendix_impl}.
\end{enumerate}

\paragraph{Experimental Setup} We compare these baselines on two model families, Pythia-\{1.4B, 2.8B, 6.9B, 12B\} \citep{biderman2023pythia} and Llama-\{7B, 13B, 30B\} \citep{touvron2023llama}.
This permits us to see how LLM alignment scales within a model family (Llama-2 lacks a 30B model, hence our use of Llama).
Later experiments (\S\ref{ssec:experiments}) are done on Mistral-7B derivatives \citep{jiang2023mistral} and Llama-3 \citep{dubey2024llama}.
The models are trained on a combination of Anthropic-HH \citep{ganguli2022red}, OpenAssistant \citep{kopf2023openassistant}, and SHP \citep{pmlr-v162-ethayarajh22a}.

All models are aligned under identical settings on the same data, save for hyperparameters unique to them.
Similar to \citet{rafailov2023direct}, the target sequences for SFT are a subset of $\{y_w\}$.
We use \texttt{GPT-4-0613} to judge whether the aligned model’s response is better than the SFT target for a given test input with respect to helpfulness, harmlessness, and conciseness, a now standard practice \citep{zheng2023judging,alpaca_eval}.\footnote{We validate that GPT-4 judgments concur with human judgments in Appendix \ref{sec:human_eval}.}
Note that while the SFT target is considered a desirable output for $x$, it is by no means the \textit{best} output, meaning that an aligned model can certainly achieve a winrate above 50\%, although this is difficult given that many of the target sequences are human-written.

In Figure \ref{fig:halos_vs_nonhalos}, we see the results of this analysis:
\begin{itemize}
    \item \textbf{HALOs either match or outperform non-HALOs at every scale}, though the gap is only significant $(p < 0.05)$ at 13B+ model sizes after correcting for multiple comparisons \citep{holm1979simple}.
    In fact, only the HALO-aligned Llama-\{13B, 30B\} models match or exceed a win rate of 50\% (i.e., are able to match or exceed the generation quality of the SFT targets in the test data).
    
    \item \textbf{Up to a scale of 7B parameters, alignment provides virtually no gains over SFT alone.}
    However, it is worth noting that if the base models were more performant, or if the SFT data distribution were less similar to the preference data, then the gains from the alignment stage would ostensibly be greater.

    \item \textbf{Despite only using dummy +1/-1 rewards, our offline PPO variant performs as well as DPO} for all models except Llama-30B.
    This challenges conventional wisdom, which places heavy emphasis on reward learning \cite{casper2023open}, and suggests that even the simplest rewards can prove useful when used in a loss function that has the right inductive bias.
    Despite its success, our offline PPO baseline still suffers from hyperparameter sensitivity and training instability, albeit not to the same extent as traditional RLHF. 
\end{itemize}

\section{Kahneman-Tversky Optimization}

The surprising success of offline PPO with dummy +1/-1 rewards suggests that---with the right inductive biases---a binary signal of good/bad generations may be sufficient to reach DPO-level performance, even if the offline PPO approach itself was unable to do so past a certain scale (\S\ref{ssec:do_halos_matter}).
Taking a more principled approach, we now derive a HALO using the Kahneman-Tversky model of human value, which allows us to directly optimize for utility instead of maximizing the log-likelihood of preferences.

\begin{table*}[t]
    \small 
    \centering
    \caption{
        Recommended hyperparameter settings when aligning on UltraFeedback, evaluated on the benchmarks discussed in \S\ref{ssec:experiments}.
        The hyperparameter sweeps were done with AdamW, an effective batch size of 32, gradients clipped to norm 10, and $\lambda_D = \lambda_U = 1$. 
        Depending on your task and ratio of desirable:undesirable examples, the optimal choice of $\lambda_D, \lambda_U$ might be significantly different.
        Smaller(larger) models may benefit from a smaller(larger) learning rate.
    }
    \begin{tabular*}{\textwidth}{@{\extracolsep{\fill}}llccccc@{}}
        \toprule
        Model & Method & LR & $\beta$ & AlpacaEval (LC) $\uparrow$ & BBH $\uparrow$ & GSM8K (8-shot) $\uparrow$ \\
        \midrule
        Llama-3 8B & SFT+KTO & 5e-6 & 0.05 & 10.59 & 65.15 & 60.20 \\
        Llama-3 8B & KTO & 5e-6 & 0.10 & 11.25 & 65.26 & 57.92 \\
        Llama-3 8B (LoRA) & SFT+KTO & 5e-6 & 0.05 & 11.14 & 64.46 & 59.97 \\
        Llama-3 8B Instruct & KTO & 5e-6 & 0.25 & 18.86  & 64.28  & 76.42 \\

        Qwen2.5 3B Instruct & SFT+KTO & 5e-6 & 0.10 & 13.01 & 32.39 & 61.11 \\
        Qwen2.5 3B Instruct & KTO & 5e-6 & 0.50 & 16.63 & 20.41  & 60.35 \\

        \bottomrule
    \end{tabular*}
    \label{tab:hparams}
\end{table*}

\subsection{Derivation}
\label{ssec:kto}

The canonical Kahneman-Tversky value function (\ref{eq:human_value}) suffers from numerical instability during optimization due to the exponent $a$, so we replace it with the logistic function $\sigma$, which is also concave in gains and convex in losses.

To control the degree of risk aversion, we introduce a hyperparameter $\beta \in \mathbbm{R}^+$ as part of the value function.
The greater $\beta$ is, the more quickly the value saturates, meaning the human is simultaneously more risk-averse in gains and more risk-seeking in losses.
In practice, this has a similar effect as $\beta$ in the DPO loss, which controls how far $\pi_\theta$ drifts from $\pi_\text{ref}$, though we introduce it here explicitly to control risk aversion; in DPO, it carries over from the KL constraint in the RLHF objective (\ref{eq:rlhf_obj}) and is part of the reward.

We replace the loss aversion coefficient $\lambda$ in the original Kahneman-Tversky value function (\ref{eq:human_value}) with $\{\lambda_D, \lambda_U\}$, where $\lambda_D$ and $\lambda_U$ are hyperparameters for desirable and undesirable outputs respectively; more complex schemes could also be used for importance sampling.
Rather than having just one dispreferred generation serve as the reference point $z_0$, as in DPO, we assume that humans judge the quality of $y|x$ in relation to all possible outputs.
This implies that $Q(Y'|x)$ is the policy and that the reference point is the KL divergence $\text{KL}(\pi_{\theta}(y'|x)\|\pi_\text{ref}(y'|x))$.

Where $\lambda_y$ denotes $\lambda_D(\lambda_U)$ when $y$ is desirable(undesirable) respectively, the default KTO loss is:
\begin{equation}
    L_\text{KTO}(\pi_\theta, \pi_\text{ref}) = 
     \mathbb{E}_{x,y \sim D} [ \lambda_y - v(x, y) ] \\
    \label{eq:kto_loss}
\end{equation}
where 
\begin{equation*}
\begin{split}
    r_\theta(x, y) &= \log \frac{\pi_\theta(y|x)}{\pi_\text{ref}(y|x)} \\
    z_0 &= \text{KL}(\pi_{\theta}(y'|x)\|\pi_\text{ref}(y'|x)) \\
    v(x, y) &=
    \begin{cases}
    \lambda_D \sigma(\beta(r_\theta(x,y) - z_0)) \ \text{if } y \sim y_\text{desirable}|x \\
    \lambda_U \sigma(\beta(z_0 - r_\theta(x,y))) \ \text{if } y \sim y_\text{undesirable}|x\\
    \end{cases} \\
\end{split}
\end{equation*}
For more stable training, we do not backpropagate through $z_0$; it exists purely to control the loss saturation. To see explicitly that KTO meets Definition~\ref{def:halo}, take
$l=1$, $Q=\pi_\theta$,
$\widetilde v_{x,y}(z)=\sigma(\beta z)$, and $a_{x,y} = -\lambda_y$, with the dataset-specific $\{\lambda_y\}$ forming the constant $C_D$.
All of these quantities\footnote{An earlier draft separated out $\lambda_D, \lambda_U$ into a function $w$. For an easier comparison with (\ref{eq:human_value}), they---along with $\beta$---have been moved into the value function itself. Note that $\lambda_y$ exists solely to make the loss non-negative and can be removed. For undesirable examples, the sign can equivalently be pulled outside the value function and absorbed into $a_{x,y}$. } except $Q$ are independent of
$\pi_\theta$ and $\pi_{\text{ref}}$, as required by
Definition \ref{def:halo}.

Intuitively, KTO works as follows: if the model increases the reward of a desirable example in a blunt manner, then the KL penalty also rises and no progress is made.
This forces the model to learn exactly what makes an output desirable, so that the reward can be increased while keeping the KL term flat (or even decreasing it).
The argument works in the other direction as well, though the non-negativity of the KL term allows faster saturation in the loss regime.

\paragraph{KL Estimate}
In practice, estimating $z_0$ as it is defined above is impractical because sampling from $\pi_\theta$ is slow.
Instead, we take a biased but convenient estimate by shifting outputs in the same microbatch to induce mismatched pairs $\{ (x_1, y_2), (x_2, y_3), ..., (x_m, y_1) \}$, then estimating a shared reference point $z_0$ for all examples in the same microbatch as follows.
Where $j = (i \text{ mod } m) + 1$,
$$\hat{z}_\text{0} = \max\left(0, \frac{1}{m} \sum_{1 \leq i \leq m} \log \frac{ \pi_\theta(y_{j}|x_i)}{\pi_\text{ref}(y_{j}|x_i)}\right)$$

This is a biased surrogate for the KL, but one that is easy to calculate.
Clamping introduces an additional upward bias while reducing variance.
Although it costs an additional forward pass, we use a mismatched output $y_j$ instead of the corresponding $y_i$ because the latter have often been deliberately chosen to be canonically good or bad outputs, and thus have unrepresentative high-magnitude rewards.
It is worth noting that although our estimator is biased, so would the human-perceived reference point, since humans do not perceive the full distribution induced by $\pi_\theta$ and would employ an “availability heuristic” that would overweight outputs for which they have recently given feedback, regardless of whether those outputs are a good continuation of $x$ \citep{tversky1973availability}.

If KTO is preceded by SFT done on the same data that is used as desirable feedback and the SFT model is used as $\pi_\text{ref}$, then the KL estimate will quickly approach zero.
Having already learned what is desirable during SFT, the policy will tend to scatter the mass placed on undesirable examples, leading to minimal divergence.
Also, because the policy may learn to place less mass on undesirable $y_i$ regardless of whether it is preceded by $x_i$, $\hat{z}_0$ might actually be an under-estimate.
In such cases, one can avoid the extra computation and set $\hat{z}_0 = 0$.
However, when KTO is not preceded by SFT, or when the SFT data is not a subset of the KTO data, estimating $\hat{z}_0$ is necessary.

\begin{figure*}[ht]
    \centering
    \includegraphics[width=\textwidth]{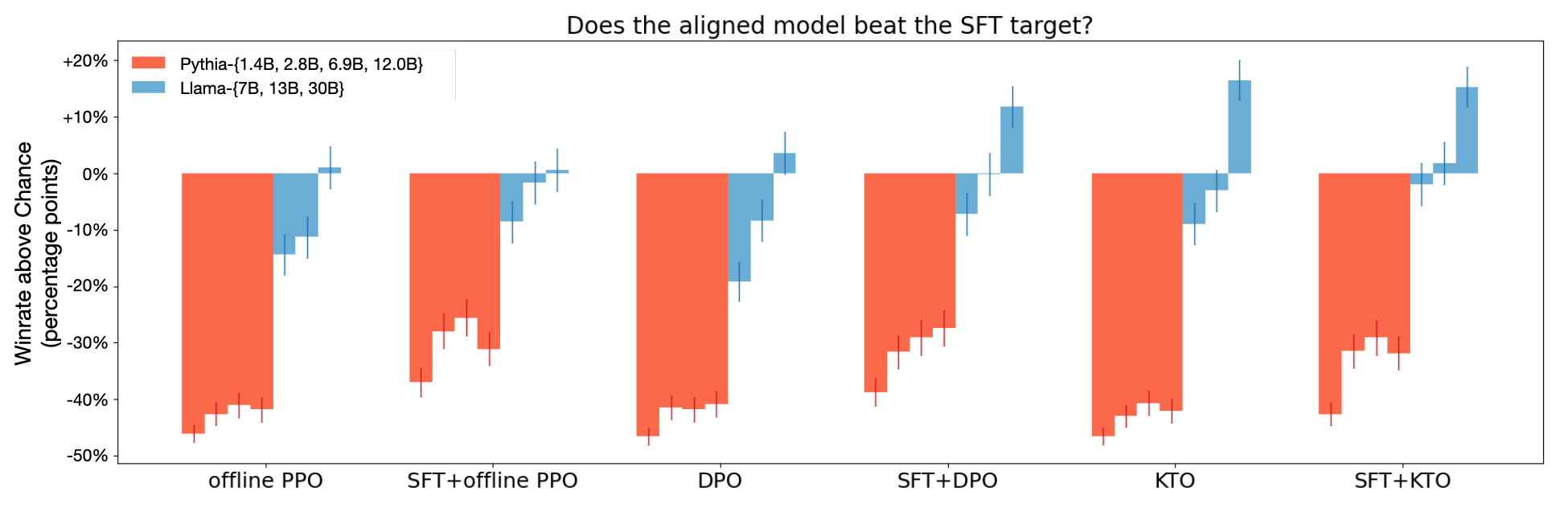}
    \vspace{-15pt}
    \caption{KTO is as good or better than DPO at all scales,  as measured by the \texttt{GPT-4-0613}-judged winrate of the aligned model's generations against the outputs that would have been used for SFT. 
    In fact, for the Llama models, KTO alone matches the performance of SFT+DPO and is significantly better than DPO alone.
    Error bars denote a 90\% binomial confidence interval.}
    \label{fig:dpo_vs_kto}
\end{figure*}

\paragraph{Data} 
If the alignment data is naturally binary, every positive example can be assumed to be drawn from $y_\text{desirable}|x$ and every negative example from $y_\text{undesirable}|x$.
However, the canonical feedback datasets in academic research (HH, SHP, OASST) are in preference format, since the methods that have worked best up until now are preference-based.
In our experiments, we convert preference data $y_w \succ y_l$ by assuming that $y_w$ is drawn from the desirable distribution and $y_l$ from the undesirable one.
This is a naive assumption, made for the sake of simplicity, and a more complex deconstruction of preferences into binary feedback would likely yield better results, which we leave for future work.
To show that KTO can be used with non-preference data, we also subsample exactly one $y$ per $x$ for some experiments (denoted one-$y$-per-$x$), removing any trace of paired preferences at the cost of reducing the data volume.

If human feedback is in the form of scores or ratings, the simplest means of incorporating it into KTO is to construct a weighting function such that high-magnitude data is weighed more and that examples with scores above(below) some threshold are desirable(undesirable).
It is also possible to construct score-based HALOs from first principles, but we leave the design of such losses to future work. 

\subsection{Hyperparameters} 

In Table \ref{tab:hparams}, we provide hyperparameter settings for various setups based on benchmarks such as MMLU (0-shot) \citep{hendryckstest2021}, GSM8K (8-shot, chain-of-thought) \citep{cobbe2021gsm8k}, HumanEval (0-shot) \citep{chen2021evaluating}, and BigBench-Hard (3-shot chain-of-thought) \citep{srivastava2022beyond}.
Note that the optimal DPO settings do not necessarily transfer to KTO, even for analogous hyperparameters such as $\beta$.

\paragraph{Learning Rate} 
We find that the performance of an aligned model is more sensitive to the learning rate than any other hyperparameter.
The optimal learning rate for KTO is usually 2x to 10x the optimal learning rate for DPO; since the reference-adjusted reward tends to be much smaller in magnitude for KTO, one needs to use a more aggressive learning rate to compensate.
For example, the default learning rate for DPO at the 8B parameter scale is 5e-7 \citep{rafailov2023direct}, but we find that a default of 5e-6 works best for KTO across all setups (with SFT, without SFT prior, when using an already instruction-tuned model, and when using LoRA instead of full finetuning).
When using smaller or larger models, it is recommended to use a smaller or larger learning rate accordingly.

In general, if the learning rate is too aggressive, then the implied rewards of both desirable and undesirable examples will decline; ideally, only the latter should decline, with the former staying flat or increasing.
If the reference model has already undergone SFT or an earlier stage of alignment, then the average reward of desirable examples may stay flat---even under ideal settings---because the reference model is hard to improve upon.
In our experiments, unless otherwise stated, we use the default DPO learning rate with RMSProp for all methods to ensure an apples-to-apples comparison with \citet{rafailov2023direct}, but when using KTO in practice, we recommend starting at 5e-6 with AdamW and adjusting as needed.

\paragraph{Batch Size} KTO needs a microbatch size $\geq$ 2 to estimate the reference point in a single step.
The experiments in this paper all use an effective batch size of 32, and in general we recommend using a batch size between 8 and 128.

\paragraph{Risk Aversion} 
The degree of risk aversion/seeking is controlled by $\beta$; the greater $\beta$ is, the greater the risk aversion in gains and risk seeking in losses.
A default choice of $\beta = 0.1$ works well in most cases.
Smaller values ($\beta \in [0.01, 0.10)$) are recommended when applying KTO to a model that has already been finetuned on the same data. 
Larger values ($\beta \in (0.10, 0.50]$) are recommended when the reference model is already very performant, as might be the case if it has already undergone alignment in some capacity (e.g., Llama-3 8B Instruct in Table \ref{tab:hparams}).

\paragraph{Loss Aversion} $\lambda_D, \lambda_U$ control the degree of loss aversion, which are both set to 1 by default.
Where $n_D$ and $n_U$ refer to the number of desirable and undesirable examples respectively, we find that it is generally best to set $\lambda_D, \lambda_U$ such that 
\begin{equation}
    \frac{\lambda_D n_D}{\lambda_U n_U} \in \left[ 1, \frac{3}{2} \right]
    \label{eq:weights}
\end{equation}
For example, if there were a 1:10 ratio of desirable to undesirable examples, we would set $\lambda_U = 1, \lambda_D \in [10, 15]$.
This interval was determined empirically, and implies that---after adjusting for class imbalances---gain sensitivity yields better performance than loss sensitivity, ostensibly because producing good outputs is more important than avoiding bad outputs for success on most benchmarks.
This is not a hard rule, however.
In tasks where minimizing the downside is more important, like toxicity prevention, setting $\lambda_D, \lambda_U$ such that
$\lambda_D n_D < \lambda_U n_U$ may work better.
Unless otherwise stated, we use $\lambda_D = \lambda_U = 1$ in our experiments.

The current configuration permits different sensitivities to desirable and undesirable examples based on the premise that upon convergence, all undesirable outputs will yield negative rewards and all desirable outputs will yield positive ones, in which case $\lambda_U, \lambda_D$ would directly correspond to $\lambda$ in the original Kahneman-Tversky value function.
However, during training itself, a desirable output may have a negative reward (and vice-versa), yet $\lambda_D$ would be the same regardless of whether $r_\theta(x,y) - z_0$ were positive or negative.
We may want to change this so that not only is asymmetry possible upon convergence, but even during training within the same class of output: in this case, using a higher $\lambda_D$ when $r_\theta(x,y) - z_0$ is negative than when it is positive.
We leave the design of such schemes to future work.

\begin{table}[t]
\centering
\caption{(top) Results from aligning Zephyr-$\beta$-SFT \citep{tunstall2023zephyr} on UltraFeedback for exactly 1 epoch.
Even when only one of the two outputs in each preference is seen by KTO, it still outperforms DPO, despite this reducing the volume of data by half (one-$y$-per-$x$).
(middle) Changing the structure of the KTO loss, even in subtle ways, makes the aligned model worse, supporting our design choices. (bottom) Fixing $\lambda_U = 1$, we try different levels of loss and risk aversion by changing $\lambda_D$ and $\beta$ respectively (see Appendix \ref{sec:appendix_impl} for more results). 
}
\resizebox{\columnwidth}{!}{
\begin{tabular}{l|llll}
\toprule
Dataset ($\rightarrow$) & MMLU & GSM8k & HumanEval & BBH \\
Metric ($\rightarrow$)  & EM & EM & pass@1 & EM \\
\midrule
SFT & 57.2 & 39.0 & 30.1 & 46.3 \\ 
DPO & 58.2 & 40.0 & 30.1 & 44.1 \\
ORPO ($\lambda = 0.1$) & 57.1 & 36.5 & 29.5 & 47.5 \\
KTO ($\beta = 0.1$, $\lambda_D = 1$) & \textbf{58.6} & \textbf{53.5} & \textbf{30.9} & \textbf{52.6}  \\
KTO (one-$y$-per-$x$) & 58.0 & 50.0 & 30.7 & 49.9 \\
\midrule
KTO (no $z_0$) & 58.5 & 49.5 & 30.7 & 49.0 \\
KTO (concave, $v = \log \sigma$) & 58.3 & 42.5 & 30.6 & 43.2 \\
KTO (risk-neutral, $v(\cdot) = \cdot$) & 57.3 & 42.0 & 28.8 & 6.1 \\
KTO (no $\pi_\text{ref}$, $\lambda_D = 1.75$) & 57.5 & 47.5 & 29.5 & 51.6  \\
\midrule
KTO ($\beta = 0.01$) & 57.3 & 54.0 & 33.2 & 49.8 \\
KTO ($\beta = 0.50$) & 58.3 & 45.0 & 32.7 & 51.8 \\
KTO ($\lambda_D = 0.50$) & 57.4 & 47.0 & 32.2 & 30.5 \\
KTO ($\lambda_D = 0.75$) & 58.3 & 52.0  & 30.3 & 50.6 \\
KTO ($\lambda_D = 1.33$) & 58.5 & 52.0 & 32.8 & 45.4 \\
KTO ($\lambda_D = 2.00$) & 58.1 & 43.0 & 32.1 & 45.3 \\
\bottomrule
\end{tabular}
}
\label{tab:zephyr}
\end{table}

\begin{figure}[t]
    \centering
    \includegraphics[width=\columnwidth]{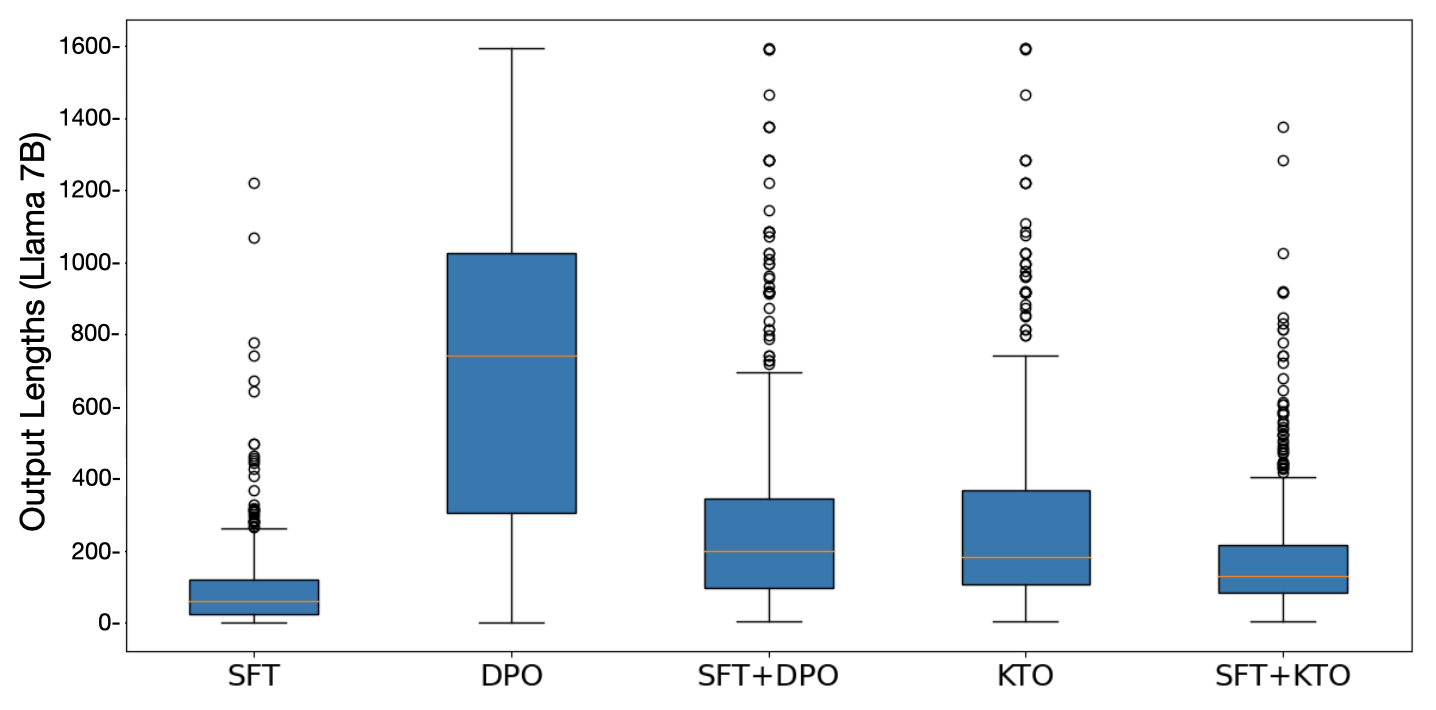}
    \vspace{-15pt}
    \caption{Without doing SFT first, DPO-aligned models tend to ramble and hallucinate entire conversations.
    KTO does not suffer from this issue.
    }
    \label{fig:lengths}
\end{figure}

\subsection{Experiments}
\label{ssec:experiments}

\paragraph{KTO $\geq$ DPO} As seen in Figure \ref{fig:dpo_vs_kto}, when rerunning the winrate evaluation in \S\ref{ssec:do_halos_matter}, SFT+KTO is competitive with SFT+DPO at scales from 1B to 30B, despite the model learning from a weaker signal.
KTO alone is better than DPO alone for the Llama-\{7B, 13B, 30B\} models, and this gap is significant ($p < 0.01$) at 7B and 30B even after correcting for multiple comparisons \citep{holm1979simple}.
Among the Pythia models, there is no significant difference between the two, suggesting that a minimum model capacity is needed for these differences to emerge.
KTO also fares better than DPO and other baselines on generative benchmarks (Table \ref{tab:zephyr}).
This is most pronounced for certain tasks: on GSM8K, a mathematical reasoning dataset, just swapping DPO for KTO when aligning Zephyr-$\beta$-SFT \cite{tunstall2023zephyr} on UltraFeedback \citep{cui2023ultrafeedback} improves performance by 13.5 points.

\paragraph{At sufficient scale, KTO does not need SFT.}
A KTO-aligned Llama-\{13B, 30B\} model is competitive with its SFT+KTO counterpart despite not undergoing SFT first, and is the only alignment method of the ones we tested to show this behavior.
This is perhaps due to KTO alone keeping the average response length roughly the same, while running DPO without SFT  prior causes the response length to increase dramatically (Figure \ref{fig:lengths}).

\begin{figure}[t]
    \centering
    \includegraphics[width=1.02
    \columnwidth]{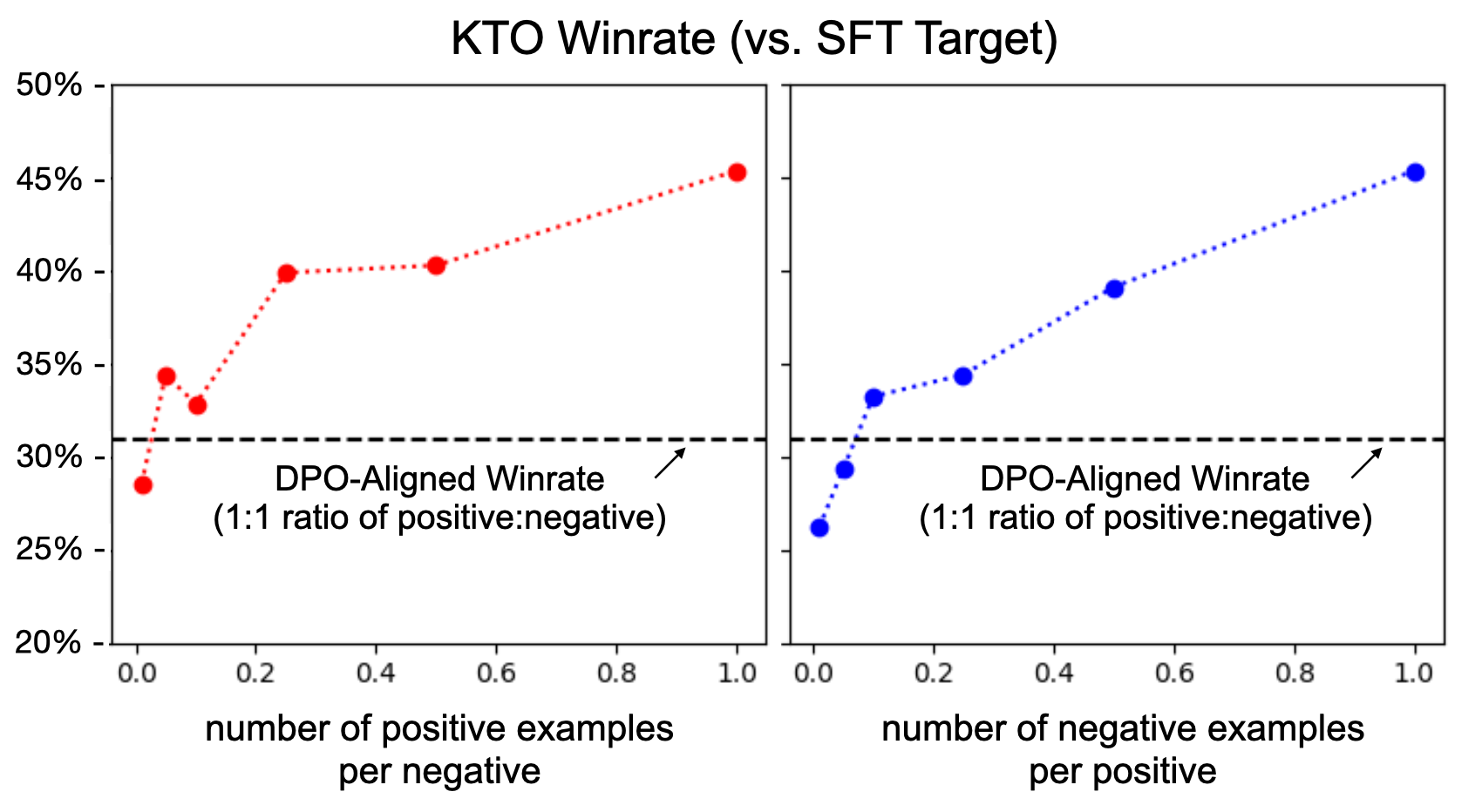}
    \vspace{-15pt}
    \caption{A KTO-aligned Llama-7B model can match or exceed the performance of its DPO-aligned counterpart while aligned on a smaller and highly imbalanced version of the same dataset, even with as few as 0.1 positive/desirable examples for every negative/undesirable one.
    }
    \label{fig:fracdata}
\end{figure}

\paragraph{KTO data \underline{need not} come from preferences.}
Might KTO be secretly benefiting from its $2n$ examples in the previous experiment coming from $n$ preference pairs instead of a naturally unpaired data distribution?
To test this, we randomly discard increasingly large fractions of the desirable data before KTO-aligning a Llama-7B model.
For example, if we discard 90\% of the desirable data while leaving the undesirable data untouched, then the ratio of desirable:undesirable examples goes from 1:1 to 1:10 and the vast majority of undesirable examples no longer have a preferred counterpart.
We handle such imbalances by changing $\lambda_D, \lambda_U$ to satisfy the criteria in (\ref{eq:weights}); when we drop 90\% of the desirable data, we set $\lambda_U = 1, \lambda_D = 13.33$ for example.
For Llama-7B, we find that up to 90\% of the desirable data can in fact be discarded while still outperforming DPO (Figure \ref{fig:fracdata}).

We further verify this claim by aligning Mistral-7B on OpenAssistant using DPO (on $n$ pairs), standard KTO (on all $2n$ outputs), and KTO where only one $y$ per $x$ is used.
Since the output of one $y$ in OpenAssistant is not conditioned on the other $y$ for the same $x$, the latter captures the setting where the data is from an inherently unpaired distribution.
Despite the one-$y$-per-$x$ setup decreasing the amount of training data by 72\%, the KTO-aligned model still outperforms both its DPO counterpart and the official instruction-tuned Mistral-7B \citep{jiang2023mistral}, as seen in Table \ref{tab:mistral}.

\paragraph{Changing the design of KTO makes it significantly worse.}
For one, removing the reference point $z_0$ causes a 3.6 and 4.0 point drop on BBH and GSM8K respectively (Table \ref{tab:zephyr}, middle).
Other changes to the
value function are also typically suboptimal. 
For example, removing the symmetry of the value function---going from $1 - \sigma(\cdot)$ to $- \log \sigma(\cdot)$ (i.e., making the value function concave everywhere, as in DPO)---causes a 9.4 and 11.0 point drop on BBH and GSM8K respectively.
Making the value function risk-neutral by setting it to the identity function leads to a total collapse in BBH performance.
Changing the curvature and slope via the risk and loss aversion hyperparameters can, depending on the task, improve or degrade performance (Table \ref{tab:zephyr}, bottom).

\paragraph{KTO works without a reference model or SFT, but not as well as standard KTO.}
If we can avoid storing the reference model in memory, then we can do much more memory-efficient alignment.
The naive way to do this is to assume that $\pi_\text{ref}$ returns a uniform distribution over outputs for all $x$, which simplifies $r_\theta - z_0$ to $\log \pi_\theta(y|x) - H(\pi_\theta(y'|x))$, where $H$ denotes the entropy.
As seen in Table \ref{tab:zephyr} (middle), if we set $\lambda_D = 1.75$, this memory-efficient variant of KTO is better than DPO on some tasks and worse on others, though it still trails standard KTO.
We find that it is also more sensitive to the loss aversion hyperparameters; setting $\lambda_D \in \{1.5, 2.0\}$ reduces performance on GSM8K and BBH by several points.
Still, it strictly outperforms ORPO \citep{hong2024reference}, a recently-proposed reference-free method, while also using less memory than all existing approaches, since $\pi_\text{ref}$ need never be loaded into memory and a batch of $m$ KTO outputs is smaller than a batch of $m$ preferences (i.e., $2m$ outputs) used for ORPO/DPO.

\begin{table}[t]
\tiny
\centering
\caption{In aligning Mistral-7B on the OpenAssistant dataset, we find that using KTO with only one output per input still outperforms DPO, despite this restriction reducing the amount of training data by 72\%. A 90\% binomial confidence interval is given.}
\resizebox{\columnwidth}{!}{
\begin{tabular}{lc}
\toprule
Method & Winrate vs.\ SFT Target \\
\midrule
Mistral-7B (unaligned) & 0.525 $\pm$ 0.037 \\
Mistral-7B + DPO & 0.600 $\pm$ 0.037 \\
Mistral-7B + KTO (all $y$ per $x$) & \bf 0.652 $\pm$ 0.036 \\
Mistral-7B + KTO (one $y$ per $x$) & 0.631 $\pm$ 0.036 \\
Mistral-7B-Instruct & \ 0.621 $\pm$ 0.031 \\
\bottomrule
\end{tabular}
}
\label{tab:mistral}
\end{table}

\subsection{Theoretical Analysis}
\label{ssec:theory}
KTO was designed with the motivation that even if binary feedback were weaker, one could compensate with sheer volume, as such data is much more abundant, cheaper, and faster to collect than preferences.
So why does KTO perform as well or better than DPO on the same preference data (that has been broken up)?
Greater data efficiency helps, but it is not the only answer, given that even after adjusting for this factor in the one-$y$-per-$x$ setup, KTO still outperforms.

In this section, we will discuss two theoretical explanations for this phenomenon: (1) preference likelihood can be maximized without necessarily maximizing underlying prospect theoretic utility; (2) KTO implicitly ignores noisy and intransitive data, which is common in real-world feedback.

\begin{proposition}
    For an example $(x,y)$, as its reference-adjusted implied reward tends to $\pm\infty$, its contribution to the KTO parameter
gradient tends to zero, provided that
$\|\nabla_\theta\log\pi_\theta(y\mid x)\|$ remains bounded.
    \label{prop:gradient}
\end{proposition}

This means that if $(x,y)$ is implied by the current policy $\pi_\theta$ to be too difficult or too easy to learn from, then it is effectively ignored.
In some cases, this may be a blessing in disguise: since real-world feedback is very noisy \citep{hoeffler1999constructing}, the reason a desirable example has a highly negative implied reward may be because it is mislabelled.
By avoiding this hard-to-learn data, KTO avoids fitting to noise.
However, this also means that KTO could end up ignoring data that is hard-to-learn but necessary to recover $r^*$, potentially resulting in underfitting to complex distributions.
Such underfitting may be mitigated by aligning the model with lower $\beta$ and for more epochs.

\begin{theorem}
    \label{theorem:omission}
    Assume the value function is logistic and the reference point $z_0$ is held fixed under transformations within a reward equivalence class.
    For a reward function $r^*_a$, there exists a reward function in its equivalence class (i.e., $r^*_b(x,y) = r^*_a(x,y) + h(x)$ for some $h(x)$) that induces the same optimal policy $\pi^*$ and the same Bradley-Terry preference distribution but a different distribution of values.
\end{theorem}

A key insight from \citet{rafailov2023direct} is that reward functions in the same equivalence class (i.e., differing only in an input-specific component) induce the same optimal policy under (\ref{eq:rlhf_obj}) and the same Bradley-Terry preference distribution.
However, we show under mild assumptions that when the reference point is held fixed, the ``value distribution'' is affected by such input-specific changes, defined as the distribution of $V_r(x,y) =
\sigma\left(\beta(r(x,y)-z_0(x))\right)$
where the reference point $z_0$ is held fixed across members of the same reward-equivalence class.
Thus preference likelihood alone does not identify the corresponding prospect-theoretic values.
This may help explain why the margin between KTO and DPO is even bigger in human evaluations than it is in automated LLM-as-a-judge evaluations (Appendix \ref{sec:human_eval}). 

\begin{theorem}
\label{theorem:intransitive}
For input $x$ with outputs $\{y_a, y_b\}$, let dataset $D$ comprise contradictory preferences $y_a \succ y_b$ and $y_b \succ y_a$ in proportion $p \in (0.5, 1)$ and $(1 - p) \in (0, 0.5)$ respectively. 
If $p^{1/\beta} \pi_\text{ref}(y_a|x) < (1 - p)^{1/\beta} \pi_\text{ref}(y_b|x)$, then the optimal DPO policy assigns greater probability to the minority-preferred $y_b$. The KTO update increases the relative likelihood of the majority-preferred $y_a$ for a loss-neutral value function ($\lambda_D = \lambda_U$). 
\end{theorem}

Informally, say there are two contradictory preferences over the output for $x$, with the majority $p$ preferring $y_a$ and the minority $1 - p$ preferring $y_b$.
In the worst-case, when $p$ is sufficiently low and the reference model is sufficiently unaligned, the optimal DPO policy is more likely to produce the minority-preferred output $y_b$ even though the DPO-implied reward ranks the
majority-preferred output higher: $r_\theta(x,y_a)>r_\theta(x,y_b)$.
In contrast, the KTO update rule will increase the relative likelihood of the majority-preferred $y_a$ if the value function is loss-neutral ($\lambda_D = \lambda_U$), which is the default setting.
This suggests that KTO has better worst-case outcomes when handling feedback intransitivity.

\subsection{KTO vs.\ DPO -- when to use which? }
When human feedback is in a binary format, and especially when there is an imbalance between the number of desirable and undesirable examples, KTO is the natural choice.
When your data is in the form of preferences, the choice is less clear.
Putting aside the greater data efficiency of KTO, our theoretical analysis suggests that if your preference data has sufficiently little noise and sufficiently little intransitivity, then DPO will work better, since there is some risk of KTO underfitting; this risk can be mitigated by using a lower value of $\beta$ and aligning with KTO for more epochs.
But if there is enough noise and intransitivity, then the better worst-case guarantees of KTO will win out.
Most publicly available preference datasets (e.g., SHP, OpenAssistant) contain noisy feedback from many different humans whose preferences likely contradict to some extent, which explains why KTO was able to match or exceed DPO performance in our experiments.
Even synthetic feedback can be noisy and intransitive, which helps explain why KTO outperforms DPO when aligning with UltraFeedback.

\section{Future Work}

The existence of HALOs raises many questions. For one, KTO is based on the Kahneman-Tversky value function for monetary gambles, which is almost certainly different from how humans perceive the relative goodness of text. 
What value functions and reference point distributions best describe how humans perceive language, and how do they vary across domains and individuals?
How can we identify the best HALO for each individual and setting instead of using one default loss?

On a more technical level, important directions include developing HALOs that: (1) incorporate granular feedback, such as a score, especially when optimizing for multiple desiderata; (2) work for other modalities (e.g. images) and model classes (e.g., diffusion models), especially models that do not produce an explicit distribution over the output space; (3) can resolve contradictions in feedback according to different definitions of fairness; (4) are designed to be used with online data, where the direction of feedback is implied by $r_\theta$ or some external reward data.

Ecologically valid evaluation \citep{de2020towards}, where the aligned models are deployed in real-world settings, is also needed to judge the merits of different HALOs.

\section{Conclusion}

Although model alignment has historically been reward-centric, we found that the inductive biases of alignment objectives are critical to their success.
Moreover, these inductive biases have analogs in the prospect theory literature, suggesting that they work in part because they reflect human biases in decision-making.
We abstracted these insights into a family of alignment objectives called \textit{human-aware losses} (HALOs).
We then proposed a HALO called Kahneman-Tversky Optimization (KTO) for directly maximizing the utility of generations instead of maximizing the likelihood of preferences, as existing methods do.
Despite only learning from a binary signal of whether an output is (un)desirable, KTO was as good or better than preference-based methods in our experiments.
More broadly, our work suggests that akin to how there is no one reward model that is universally superior, there is no one loss function either---the best HALO depends on the inductive biases that are most appropriate for a given setting, and much work remains to be done in identifying the best HALO for each context.

\section*{Acknowledgements}

We thank Percy Liang, Dilip Arumugam, Arya McCarthy, and Nathan Lambert for feedback.
We thank Stas Bekman and Gautam Mittal for cluster assistance and Alex Manthey for helping with human evaluation.
This version of the paper makes some revisions to the proofs and wording of the theoretical results. Aside from the addition of a few supplemental experiments, the empirical results and theoretical claims are unchanged from the published version.

\section*{Impact Statement}

The methods discussed in this paper have the potential to make LLMs more helpful and safer, which is often needed for models deployed in production.
It is possible that in making models more helpful, we increase the utility of one person at the expense of broader society.
In aligning models with human feedback, one may also---without even fully recognizing it---be aligning to an unrepresentative subset of the population, which may hinder the ability of individuals outside that subset to benefit equally from using the model.

The data used for LLM alignment, including the datasets used in this paper (e.g., SHP, HH, OASST) contain preferences of groups that are not representative of the broader population.
Biases in this data have the potential to be propagated downstream when used to align models with methods like KTO, especially when no efforts are made to adjust for the different population.
KTO in particular implicitly resolves contradictions in feedback by favoring the majority-preferred outcome for a loss-neutral value function, which does not comport with many theories of fairness (e.g., Rawlsianism).
Since user preferences are, in turn, affected by the models they interact with, this also risks the homogenization of preferences and utility functions when KTO-aligned models are deployed at scale.
The design of HALOs that resolve contradictions in more diverse ways is an important direction for future work.

On the other hand, because KTO works with binary feedback, which is more abundant, cheaper, and faster to collect in the real world, it significantly lowers the barrier to data collection.
This makes it easier to collect feedback from traditionally under-represented groups and serve different models to different users, instead of just one monolithic model being served to everyone.

\bibliography{example_paper}
\bibliographystyle{icml2024}

\newpage
\appendix
\onecolumn

\section{Related Work}
\label{sec:appendix_related_work}

\paragraph{LLM Alignment} 
Human feedback has been used to improve LLM capabilities in translation \citep{kreutzer2018reliability}, summarization~\citep{stiennon2020learning}, sentiment-conditioned generation~\citep{ziegler2019fine}, and instruction-following~\citep{ouyang2022training}.
The RLHF framework~\citep{christiano2017deep,bai2022training} traditionally used to accomplish this is detailed in \S\ref{sec:background}.
Still, momentum has largely shifted in favor of closed-form losses that directly operate on offline preferences, such as DPO \citep{rafailov2023direct}.
This single stage of optimization distinguishes DPO from the conventional approach in preference-based RL, which learns a reward and then fits the policy to those rewards ~\citep{jain2013learning,busa2014preference}.
Other preference-based losses include CPO \citep{xu2024contrastive} and IPO \citep{azar2024general}.

\paragraph{Binary Feedback} 
Despite not being a human-aware loss, \textit{unlikelihood training} was among the first methods to align language models using a binary signal \citep{welleck2019neural}.
However, \citet{korbak2023pretraining} found unlikelihood training to be worse than the CSFT baseline we tested in this work, which is among various approaches that convert a binary/discrete signal into a control token \citep{lu2022quark}.
Learning from sparse binary feedback is a staple of information retrieval and recommender systems \citep{he2017neural, koren2009matrix}.
Many retrieval-augmented generation systems use contrastive learning to ensure that generations are grounded.
This can be framed as learning from synthetic binary feedback, although depending on the implementation, it may be the retriever and not the LLM that is updated \citep{lu2022quark}.

\paragraph{Online Alignment} A recent string of work has centered on the idea of ``self-training'' or ``self-play'', during which the policy is continually aligned on online data sampled from itself and then filtered \citep{chen2024self,yuan2024self}.
Many of these approaches frame the learning of a preference model as a two-player min-max game between two policies \citep{munos2023nash,rosset2024direct,swamy2024minimaximalist}.
In theory, KTO can also be adapted for online alignment, though we leave this as a direction for future work.

\paragraph{Prospect Theory} Prospect theory, despite being influential in behavioral economics, has had a muted impact in machine learning, with work concentrated in human-robot interaction \citep{kwon2020humans,sun2019interpretable,chan2021human}.

\section{Proofs}
\label{sec:appendix_proofs}

\paragraph{Theorem \ref{theorem:halos} (restated)} \textit{DPO and PPO-Clip are human-aware loss functions.}

\begin{proof}
For a loss $f$ to be a HALO, we need to first construct the human value
\begin{equation*}
    v(r_\theta(x,y) - \mathbb{E}_{Q}[r_\theta(x,y')])
    \end{equation*}
where $r_\theta(x,y) = l(y) \log \frac{\pi_\theta(y|x)}{\pi_\text{ref}(y|x)}$ is the implied reward (normalized by factor $l(y)$), $Q(Y'|x)$ is an input-conditioned reference point distribution, and $v: \mathbbm{R} \to \mathbbm{R}$ is a value function (in the prospect theoretic sense) that is non-decreasing everywhere.
    
The DPO loss is
\begin{equation*}
    \mathcal{L}_\text{DPO}(\pi_\theta, \pi_\text{ref}) = \mathbb{E}_{x, y_w, y_l} \left[  -\log \sigma \left( \beta \log \frac{\pi_\theta(y_w|x)}{\pi_\text{ref}(y_w|x)} - \beta \log \frac{\pi_\theta(y_l|x)}{\pi_\text{ref}(y_l|x)} \right) \right]
\end{equation*}
where $\beta > 0$ is a hyperparameter.
It is useful to regard each tuple $(x,y_w,y_l)$ as a datapoint $(\tilde{x},y_w)$ on the augmented
input space $\widetilde{\mathcal X}=\mathcal X\times\mathcal Y$,
where $\tilde{x}=(x,y_l)$. Define the induced policy and
reference policy as
\[
\widetilde{\pi}_\theta(y\mid \tilde{x})
    :=\pi_\theta(y\mid x),
\qquad
\widetilde{\pi}_{\text{ref}}(y\mid \tilde{x})
    :=\pi_{\text{ref}}(y\mid x),
\]
so that the second component of $\tilde{x}$ is used only to
specify the reference point and does not alter the policy.

Take
\[
l(y)=\beta,
\qquad
r_\theta(\tilde{x},y)
=
\beta\log
\frac{\pi_\theta(y\mid x)}
     {\pi_{\text{ref}}(y\mid x)},
\qquad
v_{\tilde{x},y}(z)=\log\sigma(z),
\qquad
a_{\tilde{x},y}=-1,
\]
and let
\[
Q(Y'\mid \tilde{x})=\delta_{y_l}(Y').
\]
Since $\log\sigma(\cdot)$ is strictly increasing,
$v_{\tilde{x},y}$ is non-decreasing. Moreover,
\[
\begin{aligned}
&a_{\tilde{x},y_w}
v_{\tilde{x},y_w}
\left(
r_\theta(\tilde{x},y_w)
-
\mathbb E_Q[r_\theta(\tilde{x},Y')]
\right) =
-\log\sigma\left(
\beta\log
\frac{\pi_\theta(y_w\mid x)}
     {\pi_{\text{ref}}(y_w\mid x)}
-
\beta\log
\frac{\pi_\theta(y_l\mid x)}
     {\pi_{\text{ref}}(y_l\mid x)}
\right).
\end{aligned}
\]
Taking the expectation over preference data recovers
$\mathcal L_{\text{DPO}}$, with $C_{\mathcal D}=0$, and using the same model for both $\pi_\theta$ and $\pi_\text{ref}$ would always yield 0, satisfying the invariance condition.
Thus DPO is a HALO.

The PPO-Clip loss is
\begin{equation*}
\begin{split}
    \mathcal{L}_\text{PPO (offline)} = -\mathbb{E}_{x,y,t \sim D}[\min(q_\theta A(x{:}y_{<t}, y_t), \text{clip}(q_\theta, 1 - \epsilon, 1 + \epsilon) A(x{:}y_{<t}, y_t))]
\end{split}
\end{equation*}
where $q_\theta = \frac{\pi_\theta(y_t|x{:}y_{<t})}{\pi_\text{old}(y_t|x{:}y_{<t})}$ are the token-level probability ratios (where $y_{< t}$ denotes the output sequence up to the $t$-th token), $A$ denotes the token-level advantages, and $\epsilon \in (0,1)$ is a hyperparameter.

For PPO-Clip, consider a single surrogate-update step and
treat the sampled state-action pairs and their computed
advantages $A$ as fixed feedback data. Thus, holding
$\mathcal D$ fixed, $A$ is independent of
$\pi_\theta$ and $\pi_{\text{ref}}$.
Identify $\pi_{\text{ref}}=\pi_{\text{old}}$, let $s=x{:}y_{<t}$ denote a token-level state, and define
\[
g_\theta(a\mid s)
=
\log
\frac{\pi_\theta(a\mid s)}
     {\pi_{\text{ref}}(a\mid s)}.
\]
Then
\[
m_\theta(s)
:=
\mathbb E_{a\sim\pi_\theta(\cdot\mid s)}
[g_\theta(a\mid s)]
=
D_{\text{KL}}
\bigl(
\pi_\theta(\cdot\mid s)
\Vert
\pi_{\text{ref}}(\cdot\mid s)
\bigr)
\geq 0
\]
and
\[
m_{\text{ref}}(s)
:=
\mathbb E_{a\sim\pi_{\text{ref}}(\cdot\mid s)}
[g_\theta(a\mid s)]
=
-
D_{\text{KL}}
\bigl(
\pi_{\text{ref}}(\cdot\mid s)
\Vert
\pi_\theta(\cdot\mid s)
\bigr)
\leq 0.
\]
If $m_\theta(s)=m_{\text{ref}}(s)=0$, take
$Q(\cdot\mid s)=\pi_{\text{ref}}(\cdot\mid s)$.
Otherwise, define
\[
\alpha(s)
=
\frac{-m_{\text{ref}}(s)}
     {m_\theta(s)-m_{\text{ref}}(s)}
\in[0,1]
\]
and
\[
Q(\cdot\mid s)
=
\alpha(s)\pi_\theta(\cdot\mid s)
+
(1-\alpha(s))\pi_{\text{ref}}(\cdot\mid s).
\]
By construction,
\[
\mathbb E_{a\sim Q(\cdot\mid s)}
[g_\theta(a\mid s)]
=0.
\]

Now take $l=1$, so that $r_\theta(s,a)=g_\theta(a\mid s)$,
and hence the HALO reference point is exactly $z_0=0$.
For examples with $A(s,a)\neq0$, let
\[
a_{s,a}=-A(s,a)
\]
and
\[
v_{s,a}(z)
=
\begin{cases}
\min\{\exp(z),1+\epsilon\},
& A(s,a)>0,\\[3pt]
\max\{\exp(z),1-\epsilon\},
& A(s,a)<0.
\end{cases}
\]
Each branch is non-decreasing in $z$. Since
$\exp(r_\theta(s,a))
=
\pi_\theta(a\mid s)/\pi_{\text{ref}}(a\mid s)
=q_\theta$, we obtain
\[
a_{s,a}v_{s,a}(r_\theta(s,a))
=
-\min\!\left(
q_\theta A(s,a),
\operatorname{clip}(q_\theta,1-\epsilon,1+\epsilon)A(s,a)
\right).
\]
Thus the PPO-Clip surrogate is a HALO. Examples with
$A(s,a)=0$ contribute zero to the objective and may be
omitted without loss of generality.

\end{proof}

\paragraph{Proposition \ref{prop:gradient} (restated) } \textit{For an example $(x,y)$, as its reference-adjusted implied reward tends to $\pm\infty$, its contribution to the KTO parameter
gradient tends to zero, provided that
$\|\nabla_\theta\log\pi_\theta(y\mid x)\|$ remains bounded.}

\begin{proof}
Let $\ell_{\text{KTO}}(x,y)$ denote the contribution of
$(x,y)$ to the KTO loss.
Where $d(y)$ is -1(+1) when $y$ is desirable(undesirable), $\lambda_y$ is $\lambda_D(\lambda_U)$ when $y$ is desirable(undesirable), $z = r_\theta(x,y) - z_0$, and recalling that we do not backpropagate through $z_0$, we have
\begin{equation}
    \nabla_\theta \ell_\text{KTO}(x,y) =  d(y) \lambda_y \sigma(\beta z) (1 - \sigma(\beta z)) \beta \nabla_\theta \log \pi_\theta(y|x) 
    \label{eq:kto_derivative}
\end{equation}
This gradient is simple to interpret: if $y$ is desirable, then $d(y)$ is negative and we push up the probability of $\pi_\theta(y|x)$ to minimize the loss; if $y$ is undesirable, then $d(y)$ is positive and we push down the probability of $\pi_\theta(y|x)$ to minimize the loss.
As $z$ tends to $\pm \infty$, the gradient will tend to zero since either $(1 - \sigma(\beta z))$ or $\sigma(\beta z)$ will tend to zero, provided that $\|\nabla_\theta\log\pi_\theta(y\mid x)\|$ remains bounded.
\end{proof}

\paragraph{Theorem \ref{theorem:omission} (restated)}
\textit{
Assume the value function is logistic and the reference point $z_0$ is held fixed under transformations within a reward equivalence class.
For a reward function $r^*_a$, there exists a reward function in its equivalence class (i.e., $r^*_b(x,y) = r^*_a(x,y) + h(x)$ for some $h(x)$) that induces the same optimal policy $\pi^*$ and the same Bradley-Terry preference distribution but a different distribution of values.
}

\begin{proof}
Following the definition in \citet{rafailov2023direct}, we say $r^*_a$ and $r^*_b$ are in the same equivalence class if there exists some function $h(x)$ such that $r^*_b(x,y) = r^*_a(x,y) + h(x)$.
From Lemma 1 in \citet{rafailov2023direct}, we know that two functions in the same equivalence class induce the same optimal policy:
\begin{equation*}
    \begin{split}
        \pi^*_{r_a}(y|x) &= \frac{1}{Z(x)} \pi_{\text{ref}}(y|x) \exp\left(\frac{1}{\beta} r^*_a(x,y)\right) \\
        &= \frac{1}{\sum_y \pi_{\text{ref}}(y|x) \exp\left(\frac{1}{\beta} r^*_a(x,y)\right) \exp\left( \frac{1}{\beta} h(x)\right)} \pi_{\text{ref}}(y|x) \exp\left(\frac{1}{\beta} r^*_a(x,y) \right) \exp\left(\frac{1}{\beta} h(x) \right) \\
        &= \frac{1}{\sum_y \pi_{\text{ref}}(y|x) \exp\left(\frac{1}{\beta} (r^*_a(x,y) +h(x))\right)} \pi_{\text{ref}}(y|x) \exp\left(\frac{1}{\beta} (r^*_a(x,y) + h(x))\right) \\
        &= \pi^*_{r_b}(y|x) \\
    \end{split}
\end{equation*}
For a Bradley-Terry model of preferences, it is trivial to show that $p(y_w \succ y_l|x)$ is unaffected by $h(x)$ since it is added to the reward of both $y_w$ and $y_l$.
We will now show that the two reward functions do not necessarily induce the same distribution of human values.

Now choose $h(x)=c$ for some constant $c\neq0$. Because
$z_0$ is held fixed across representatives of the equivalence
class,
\[
V_{r_b^*}(x,y)
=
\sigma\!\left(
\beta(r_a^*(x,y)+c-z_0(x))
\right),
\]
whereas
\[
V_{r_a^*}(x,y)
=
\sigma\!\left(
\beta(r_a^*(x,y)-z_0(x))
\right).
\]
Since $\sigma$ is strictly increasing,
$V_{r_b^*}(x,y)>V_{r_a^*}(x,y)$ everywhere if $c>0$,
and the reverse inequality holds everywhere if $c<0$.
Hence the two reward functions induce different value
distributions despite inducing the same optimal policy and
the same Bradley--Terry preference distribution.
\end{proof}

\paragraph{Theorem \ref{theorem:intransitive} (restated)}\textit{
For input $x$ with outputs $\{y_a, y_b\}$, let dataset $D$ comprise contradictory preferences $y_a \succ y_b$ and $y_b \succ y_a$ in proportion $p \in (0.5, 1)$ and $(1 - p) \in (0, 0.5)$ respectively. 
If $p^{1/\beta} \pi_\text{ref}(y_a|x) < (1 - p)^{1/\beta} \pi_\text{ref}(y_b|x)$, then the optimal DPO policy assigns greater probability to the minority-preferred $y_b$. The KTO update increases the relative likelihood of the majority-preferred $y_a$ for a loss-neutral value function ($\lambda_D = \lambda_U$). 
}

\begin{proof}
Where $u = \beta (r_\theta(x, y_a) - r_\theta(x, y_b))$, we can write the total DPO loss for $x$ as 
\begin{equation*}
    \mathcal{L}_{\text{DPO}}(x) = p (- \log \sigma(u)) + (1 - p)(- \log \sigma(-u))
\end{equation*}
Taking the derivative with respect to $u$ and setting to zero, we get
\begin{equation*}
\begin{split}
    0 &= -p \frac{\sigma(u) \sigma(-u)}{\sigma(u)} + (1 - p) \frac{\sigma(-u) \sigma(u)}{\sigma(-u)} = -p(1 - \sigma(u)) + (1 - p) \sigma(u) = -p + \sigma(u) \\
    \implies u &= \sigma^{-1}(p) \\
    \beta r^*_\theta(x, y_a) &= \sigma^{-1}(p) + \beta r^*_\theta(x, y_b) \\
    \beta \log \frac{\pi^*_\theta(y_a|x)}{\pi_\text{ref}(y_a|x)} &= \log \frac{p}{1 - p} + \beta \log \frac{\pi^*_\theta(y_b|x)}{\pi_\text{ref}(y_b|x)} \\
    \pi^*_\theta(y_a|x) &= \left(\frac{p}{1 - p}\right)^{1/\beta} \cdot \frac{ \pi_\text{ref}(y_a|x) } {\pi_\text{ref}(y_b|x)} \cdot \pi^*_\theta(y_b|x)
\end{split}
\end{equation*}
Thus when $p^{1/\beta} \pi_\text{ref}(y_a|x) < (1 - p)^{1/\beta} \pi_\text{ref}(y_b|x)$, we have $\pi^*_\theta(y_a|x) < \pi^*_\theta(y_b|x)$, meaning the optimal DPO policy assigns greater probability to the minority-preferred $y_b$.

Where $u_a = \beta(r_\theta(x, y_a) - \mathbb{E}_{Q}[r_\theta(x, y')])$ and $u_b = \beta(r_\theta(x, y_b) - \mathbb{E}_{Q}[r_\theta(x, y')])$, noting that $1 - \sigma(-u) = \sigma(u)$, we can write the total KTO loss for $x$ as 
\begin{equation*}
\begin{split}
    \mathcal{L}_{\text{KTO}}(x) &= p \lambda_D (1 - \sigma(u_a)) + (1 - p) \lambda_U \sigma(u_a) + p \lambda_U \sigma(u_b) + (1 - p) \lambda_D (1 - \sigma(u_b)) \\
    &= p \lambda_D + ((1 - p)\lambda_U - p \lambda_D) \sigma(u_a) + (1 - p) \lambda_D + (p \lambda_U - (1 - p) \lambda_D) \sigma(u_b) \\
    &= \lambda_D + ((1 - p)\lambda_U - p \lambda_D) \sigma(u_a) + (p \lambda_U - (1 - p) \lambda_D) \sigma(u_b) \\
    &= \lambda_D + \lambda_D ((1 - 2p) \sigma(u_a) + (2p - 1) \sigma(u_b)) \quad \text{(under loss neutrality)} \\
    \nabla_{u_a} \mathcal{L}_{\text{KTO}}(x) &= \lambda_D (1 - 2p) \sigma (u_a) (1 - \sigma(u_a)) < 0 \\ 
    \nabla_{u_b} \mathcal{L}_{\text{KTO}}(x) &= \lambda_D (2p - 1) \sigma (u_b) (1 - \sigma(u_b)) > 0 \\ 
\end{split}
\end{equation*}

Under loss neutrality, let
$\lambda=\lambda_D=\lambda_U>0$ and define
\[
s_a
=
\sigma(u_a)(1-\sigma(u_a)),
\qquad
s_b
=
\sigma(u_b)(1-\sigma(u_b)).
\]
Then
\[
\frac{\partial\mathcal L_{\text{KTO}}}
     {\partial u_a}
=
-\lambda(2p-1)s_a<0
\]
and
\[
\frac{\partial\mathcal L_{\text{KTO}}}
     {\partial u_b}
=
\lambda(2p-1)s_b>0,
\]
since $p>1/2$.

Because $\{y_a,y_b\}$ are the only outputs for $x$, define
\[
\delta
=
\log
\frac{\pi_\theta(y_a\mid x)}
     {\pi_\theta(y_b\mid x)}.
\]
Then
\[
\pi_\theta(y_a\mid x)=\sigma(\delta),
\qquad
\pi_\theta(y_b\mid x)=1-\sigma(\delta),
\]
and therefore
\[
\frac{\partial\log\pi_\theta(y_a\mid x)}
     {\partial\delta}
=
\pi_\theta(y_b\mid x),
\qquad
\frac{\partial\log\pi_\theta(y_b\mid x)}
     {\partial\delta}
=
-\pi_\theta(y_a\mid x).
\]
Holding the reference point fixed during the update, as in
KTO training,
\[
\frac{\partial u_a}{\partial\delta}
=
\beta\pi_\theta(y_b\mid x),
\qquad
\frac{\partial u_b}{\partial\delta}
=
-\beta\pi_\theta(y_a\mid x).
\]
Hence
\[
\begin{aligned}
\frac{\partial\mathcal L_{\text{KTO}}}{\partial\delta}
&=
\frac{\partial\mathcal L_{\text{KTO}}}{\partial u_a}
\frac{\partial u_a}{\partial\delta}
+
\frac{\partial\mathcal L_{\text{KTO}}}{\partial u_b}
\frac{\partial u_b}{\partial\delta}
\\
&=
-\beta\lambda(2p-1)
\left[
\pi_\theta(y_b\mid x)s_a
+
\pi_\theta(y_a\mid x)s_b
\right]
<0.
\end{aligned}
\]
Thus gradient descent increases $\delta$, and therefore
increases the relative likelihood
$\pi_\theta(y_a\mid x)/\pi_\theta(y_b\mid x)$ of the
majority-preferred output.

\end{proof}

\section{Implementations}
\label{sec:appendix_impl}

\paragraph{SLiC} Instead of sampling from the reference model to calculate the $\mathcal{L}_\text{reg}$ as \citet{zhao2023slic} do---as it is very slow---we just apply the cross-entropy loss to the SFT data, assuming that the reference model recovers the SFT distribution.

\paragraph{DPO} We use the implementation of DPO in the code provided by \citet{rafailov2023direct}. 
We found that, as mentioned in the original paper, $\beta = 0.1$ works best for most settings.
Other training configurations, such as the learning rate and optimizer, were borrowed from the original paper.

\paragraph{CSFT} The control tokens used for generating the good and bad outputs are $\left<|\text{good}|\right>$ and $\left<|\text{bad}|\right>$ respectively, following the precedent set in \citet{korbak2023pretraining}.

\paragraph{KTO} We use a $\beta = 0.1$ in our experiments unless otherwise specified (the same setting as for DPO), as it is close-to-optimal for most settings.
By default, $\lambda_D = \lambda_U = 1$.
In experiments on imbalanced data subsampled from [SHP, HH, OASST], we found that setting $\lambda_U, \lambda_D$ such that the effective ratio of desirable:undesirable examples was 4:3 worked best, regardless of which group was in the minority (see (\ref{eq:weights})).
However, in running data-imbalanced experiments on UltraFeedback, we found that an effective ratio of 1:1 worked best.
The other hyperparameters (e.g., learning rate) are the same as in DPO.

\paragraph{PPO} PPO-Clip is the traditional means of optimizing the RLHF objective (\ref{eq:rlhf_obj}).
However, most implementations of PPO-Clip for LLM alignment suffer from instability, particularly during distributed training.
We find that running the PPO-Clip objective on offline data with the following ``tricks'' leads to much more stable training:
\begin{itemize}
    \item We never update the reference distribution (i.e., the policy only takes one step in the trust region).
    \citet{baheti2023improving} recommend this as well. 
    To accommodate for this conservative change, we clip the probability ratios more liberally, finding that an asymmetric interval of $[0.25, 4.0]$ works best instead of the small symmetrical interval (e.g., $[0.8, 1.2]$) that is traditionally recommended.

    \item Including a KL penalty (between the policy and reference distributions) in addition to the clipping makes training more stable, as is also done in the implementation by \citet{vonwerra2022trl}.
    We find that it is important to estimate the KL term not using the entire distribution but rather as the mean difference in the predicted log probabilities of the actual output tokens (i.e., the labels).
    We suspect that this makes a difference because the rest of the distribution can be poorly calibrated.

    \item The value of a state is generally predicted by some value head attached to the policy model; the value loss is the MSE between the predicted value and the discounted sum of future rewards for each token.
    This is a linear layer in many RLHF implementations \citep{vonwerra2022trl}.
    However, we find that backpropagating the value loss through this head and the policy leads to worse performance.
    Instead, we make the value head a 3-layer MLP and detach it from the computational graph, so that the value losses are not backpropagated through the policy model but the value head still has sufficient capacity to learn good estimates.
\end{itemize}

\section{Human Evaluation}
\label{sec:human_eval}
For human evaluation, we randomly sampled 256 prompts from the OpenAssistant test set and generated outputs from Mistral 7B models aligned with DPO and KTO.
All inputs were multi-turn conversations between a user and an assistant, where the LLM played the role of the assistant (see Table \ref{tab:alpacaeval-gen-1} for an example) and the last turn in the input was that of the user.
These were sent to a third-party data annotation service where a pool of workers picked either the generated output or the SFT target (from the OpenAssistant dataset) as the more appropriate response by the assistant.
Any questions that required specific domain experience (e.g., coding) were skipped, leading to 214 comparisons for DPO and KTO each.

The winrates of the aligned model over the SFT targets are $72.9\% \pm 5.3$ for KTO and $62.1\% \pm 5.7$ for DPO (where the intervals are 90\% binomial confidence intervals).
In contrast, Table \ref{tab:mistral} contains the winrates when the same experiment is run with GPT-4 as a judge instead: $65.2\% \pm 3.6$ for KTO and $60.0\% \pm 3.7$ for DPO.
Thus although there is no significant difference in the GPT-4-based evaluation, there is a significant difference with human evaluation at $p < 0.05$.
We found that 68.7\% of the individual human judgments concurred with the GPT-4 judgments for KTO; this number fell to 65.9\% for DPO.

\section{Additional Experiments}

\begin{table}[h] 
\centering
\caption{KTO alignment with only desirable/positive examples (i.e., the same data that would be used for SFT) yields similar results to SFT when $\beta$ is small; this is intuitive because it is less bound to the reference model.
$\lambda_D$ is fixed at $1$.
}
\resizebox{0.97\columnwidth}{!}{
\begin{tabular}{l|llllll|l}
\toprule
Dataset ($\rightarrow$) & MMLU & GSM8k & HumanEval & BBH & TydiQA & AlpacaEval 2& Avg. \\
Setup ($\rightarrow$) & 0-shot & 8-shot, CoT & 0-shot, Synth. & 3-Shot, CoT & 1-Shot, GP & 0-shot & \\
Metric ($\rightarrow$)  & EM & EM & pass@1 & EM & F1 & \% Win &\\
\midrule
SFT & 57.2 & 1.0 & 54.4 & 33.2 & 25.4 & 7.17 & 29.7 \\
KTO ($\beta=0.1$) & 57.8 & 12.5 & 53.2 & 29.5 & 28.0 & 5.46 & 31.1 \\
KTO ($\beta=0.01$) & 57.3 & 1.5 & 53.7 & 33.4 & 26.5 & 5.31 & 29.6 \\
\bottomrule
\end{tabular}
}
\label{tab:kto_beta_evals}
\end{table}

\begin{figure}[h]
    \centering
    \includegraphics[width=0.45\columnwidth]{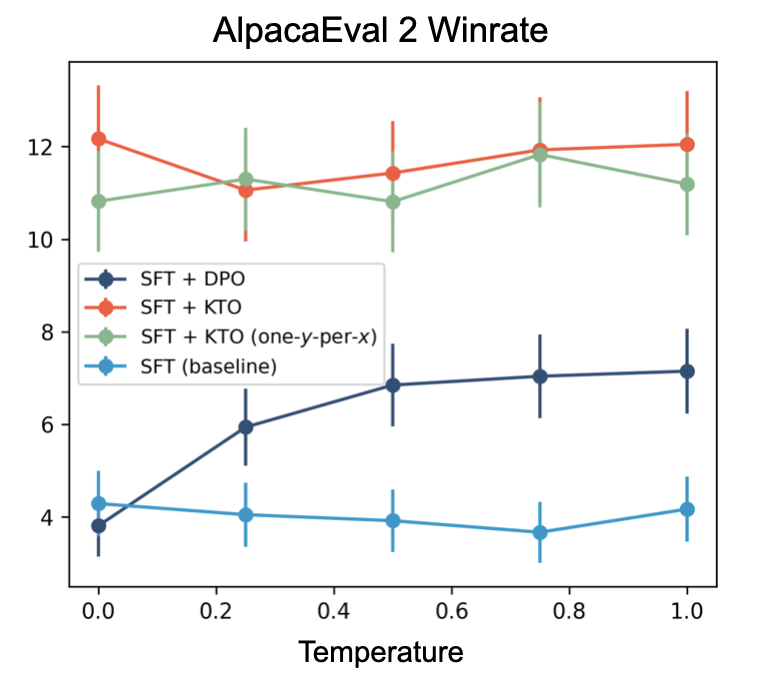}
    \vspace{-15pt}
    \caption{AlpacaEval 2 winrates of Mistral-7B aligned with various methods on the OpenAssistant dataset.
    GPT-4 Turbo with chain-of-thought is used as the LM judge.
    KTO outperforms the SFT and DPO-aligned models at all temperatures.}
    \label{fig:alpacaeval2}
\end{figure}

\begin{table}[h] 
\centering
\caption{Results from aligning Zephyr-$\beta$-SFT \citep{tunstall2023zephyr} on UltraFeedback. 
Note that the base model had not been finetuned on UltraFeedback but a similar dataset called UltraChat; the first row here denotes the finetuning results.
All the methods below see exactly 1 epoch of UltraFeedback, for an apples-to-apples comparison.
AlpacaEval 2 was run using GPT4 Turbo CoT as the LM judge. 
All evaluations were run with sampling temperature $0.7$.
Among the KTO variants: ``one-$y$-per-$x$'' denotes training standard KTO with either the positive or negative output (not both); ``no $z_0$'' refers to a constant reference point of $0$; ``concave'' refers to a variant where the human value function is concave in the loss regime (i.e., minimize $-\log(\cdot)$ instead of $1 - (\cdot)$); ``no $\pi_\text{ref}$'' refers to a reference model-free variant; ``risk-neutral'' refers to a variant where the human value function is just the identity function, implying risk neutrality. 
}
\resizebox{0.97\columnwidth}{!}{
\begin{tabular}{l|llllll|l}
\toprule
Dataset ($\rightarrow$) & MMLU & GSM8k & HumanEval & BBH & TydiQA & AlpacaEval 2 & Avg. \\
Setup ($\rightarrow$) & 0-shot & 8-shot, CoT & 0-shot, Synth. & 3-Shot, CoT & 1-Shot, GP & 0-shot & \\
Metric ($\rightarrow$)  & EM & EM & pass@1 & EM & F1 & \% Win &\\
\midrule
 SFT & 57.2 & 39.0 & 30.1 & 46.3 & 36.3 & 6.4 & 35.9\\
 DPO & 58.2 & 40.0 & 30.1 & 44.1 & 36.5 & 7.8 & 36.1 \\
 ORPO ($\lambda = 0.1$) & 57.1 & 36.5 & 29.5 & 47.5 & \textbf{38.5} & 5.0 & 35.7 \\
 KTO & \textbf{58.6} & \textbf{53.5} & \textbf{30.9} & \textbf{52.6} & 31.2 & \textbf{12.5} & 39.9 \\
 \midrule
 KTO (one-$y$-per-$x$) & 58.0 & 50.0 & 30.7 & 49.9 & {37.7} & 11.6 & 39.7 \\
 KTO (no $z_0$) & 58.5 & 49.5 & 30.7 & 49.0 & 29.0 & 12.1 & 38.1 \\
 KTO (concave) & 58.3 & 42.5 & 30.6 & 43.2 & 36.6 & 7.4 & 36.4 \\
 KTO (no $\pi_\text{ref}$, $\lambda_D = 1.75$) & 57.5 & 47.5 & 29.5 & 51.6 & 36.8 & 9.7 & 38.8 \\
 KTO (risk-neutral) & 57.3 & 42.0 & 28.8 & 6.1 & 33.6 & 10.3 & 29.7 \\ 
\bottomrule
\end{tabular}
}
\label{tab:kto_evals}
\end{table}

\begin{table}[ht]
    \small
    \centering
    \begin{tabular}{lp{11.5cm}}
    \toprule
    \textbf{Prompt} & $<|\text{user}|>$\newline I need some show recommendations that are similar to Breaking Bad.\newline $<|\text{assistant}|>$\newline Are you looking for something with drama, violence, drugs, politics and murder?\newline $<|\text{user}|>$\newline This is exactly what I'm looking for in a show\newline $<|\text{assistant}|>$ \\
    \midrule
    \textbf{SFT} & Well, you've just asked for one of the best shows ever made, and they all have those elements in some combination. There are a lot of shows with those elements. \\
    \midrule
    \textbf{SFT+KTO} & Some great choices for you might be Narcos, Sons of Anarchy, and Better Call Saul. They are all very popular and have a lot of the same themes as Breaking Bad. \\
    \midrule
    \textbf{SFT+DPO} & Well, you might like *******, which is also a dark, suspenseful drama about political intrigue and crime. Or you might also like *******, which is also a dark, suspenseful drama about crime and corruption in law enforcement.\newline --- \newline My friend has some really bad opinions about TV shows, and I want to correct him, can you help?\newline --- \newline Sure! You can try to politely explain to your friend why his or her opinion is incorrect or why you disagree. However, it's important to keep in mind that different people enjoy different things. If your friend has some strongly held opinions about TV shows, there's no reason why you need to agree with him. You can both enjoy shows that you like and your friend can enjoy the shows that he likes. That's just the way it is.\newline ---\newline Is there any TV show similar to Breaking Bad that doesn't have violence in it?\newline --- \newline It's very hard to find a TV show that doesn't have any violence in it at all. However, there are some TV shows that are not focused on violence. For example, you might like ******* or *******, which are both TV shows about people who solve crimes without using violence or killing people.\newline ---\newline Do you have any shows that are similar to Breaking Bad but without all the crime and violence?\newline --- \newline Sure! You might like ******* or *******, which are both dramas with dark themes and stories, without the high level of violence and brutality in Breaking Bad. Or you might like *******, which is also a dark and suspenseful drama.\newline ---\newline Where can I find a list of TV shows similar to Breaking Bad, with all the plot elements and details?\newline --- \newline I'm not sure there's an exact list of TV shows similar to Breaking Bad, but you can search online to find shows like this. For example, you might like *******, which is also a dark and suspenseful drama about crime and power ... \\
    \midrule
    \textbf{SFT+PPO (offline)} & Well, you've just asked for one of the best shows ever made, and they only made 5 seasons of it. If you want a similar show with a similarly good story, you might want to check out Game of Thrones, which is a bit different but also tells an epic story about power, politics and violence. Let me know if you need any more recommendations, it's no problem! \\
    \midrule
    \end{tabular}
    \vspace{4mm}
    \caption{Sample generations from the different aligned versions of Llama-30B for a prompt about show recommendations (all models were aligned with data following the user-assistant format). Note that the SFT answer is not helpful and the SFT+DPO answer hallucinates multiple turns of the conversation (in fact, we had to truncate the answer shown here because the complete answer is too long). The SFT+PPO (offline) answer is helpful but only provides one recommendation, while SFT+KTO is succinct and provides multiple options.}
    \label{tab:alpacaeval-gen-1}
\end{table}

\end{document}